\documentclass[preprint,aps,pre,letterpaper,onecolumn,superscriptaddress]{revtex4-2} 

\usepackage{algorithm}
\usepackage{algpseudocode}
\usepackage[margin=1in]{geometry}
\usepackage{graphicx}
\usepackage{amssymb}
\usepackage{amsmath}
\usepackage{mathrsfs}
\usepackage{placeins}
\usepackage{caption3}
\usepackage{wrapfig}
\usepackage{caption} 
\usepackage{subcaption}
\usepackage{xcolor}
\usepackage{contour}
\usepackage{color}
\usepackage{outlines}
\usepackage[bookmarksnumbered=true,breaklinks=true,colorlinks,citecolor=blue,linkcolor=blue]{hyperref}
\usepackage{soul} 
\usepackage{comment} 
\usepackage[normalem]{ulem}
\usepackage[font=small,labelfont=bf,
justification=raggedright,
format=plain]{caption} 

\newcommand{\IM}{\mathcal{M}}
\newcommand{\ind}{\mathcal{I}}

\newcommand{\mbf}[1]{\boldsymbol{#1}}

\newcommand{\reals}{\mathbb{R}} 

\newcommand{\by}{\mbf{y}}

\newcommand{\bu}{\mbf{u}}
\newcommand{\bq}{\mbf{q}}

\newcommand{\grad}{\mbf{\nabla}} 
\newcommand{\Rey}{\mathrm{Re}}

\newcommand{\abs}[1]{\left|#1\right|}

\newcommand{\sgn}{\textrm{sgn}}

\newlength{\figwidth}
\newlength{\SCwidth}
\def\XXint#1#2#3{{\setbox0=\hbox{$#1{#2#3}{\int}$}
		\vcenter{\hbox{$#2#3$}}\kern-.5\wd0}}

\definecolor{ao(english)}{rgb}{0.0, 0.5, 0.0}

\newcommand{\MDGrevise}[1]{\textcolor{black}{#1}}

\newcommand{\ALrevise}[1]{\textcolor{black}{#1}}
\newcommand{\CEP}[1]{\textcolor{black}{#1}}
\begin{document}


\title{Building symmetries into data-driven manifold dynamics models for complex flows: application to two-dimensional Kolmogorov flow}




\author{Carlos E. P\'erez De Jes\'us}
\affiliation{Department of Chemical and Biological Engineering, University of Wisconsin-Madison, Madison WI 53706, USA}

\author{Alec J. Linot}
\affiliation{Department of Mechanical and Aerospace Engineering, University of California-Los Angeles, Los Angeles CA 90095, USA}

\author{Michael D. Graham}
\email{mdgraham@wisc.edu}
\affiliation{Department of Chemical and Biological Engineering, University of Wisconsin-Madison, Madison WI 53706, USA}

\date{\today}

\begin{abstract}

Data-driven reduced-order models of the dynamics of complex flows are important for tasks related to design, understanding, prediction, and control. Many flows obey symmetries, and the present work illustrates how these can be exploited to yield highly efficient low-dimensional data-driven models for chaotic flows. In particular, incorporating symmetries both guarantees that the reduced order model automatically respects them and dramatically increases the effective density of data sampling. Given data for the long-time dynamics of a system, and knowing the set of continuous and discrete symmetries it obeys, the first step in the methodology is to identify a ``fundamental chart'', a region in the state space of the flow to which all other regions can be mapped by a symmetry operation, and a set of criteria indicating what mapping takes each point in state space into that chart.   We then find a low-dimensional coordinate representation of the data in the fundamental chart with the use of an autoencoder architecture that also provides an estimate of the dimension of the invariant manifold where data lie. Finally, we learn dynamics on this manifold with the use of neural ordinary differential equations. We apply this method, denoted  ``symmetry charting" to simulation data from two-dimensional Kolmogorov flow in a chaotic bursting regime. This system has a continuous translation symmetry,  and discrete rotation and shift-reflect symmetries. With this framework we observe that less data is needed to learn accurate data-driven models, more robust estimates of the {manifold dimension} are obtained, equivariance of the NSE is satisfied, better short-time tracking with respect to the true data is observed, and long-time statistics are correctly captured.

\end{abstract}


\maketitle

\section{Introduction}

In recent years, neural networks (NNs) have been implemented to learn data-driven {low-dimensional} representations and dynamical models of flow problems, with  success in systems including  the Moehlis-Faisst-Eckhardt (MFE) model \cite{srinivasan2019predictions}, Kolmogorov flow \cite{Page2020, doan2021auto}, and minimal turbulent channel flow \cite{nakamura2021convolutional}.
For the most part however, these approaches do not explicitly take advantage of the fact that for dissipative systems like the Navier-Stokes Equations (NSE), the long-time dynamics are expected to lie on an invariant manifold $\mathcal{M}$ (sometimes called an inertial manifold \cite{foias1988modelling, temam1989inertial, zelik2022attractors}), whose dimension  $d_\IM$ may be much smaller than the nominal number of degrees of freedom $N$ required to specify the state of the system.  {Ideally, one could identify $d_\IM$ from data, find a coordinate representation for points on $\IM$, and learn the time-evolution on $\IM$ in those coordinates.  This would be a minimal-dimensional data-driven dynamic model.} 
 Linot \& Graham explored this approach for chaotic dynamics of the Kuramoto-Sivashinsky equation (KSE) \cite{linot2020deep,linot2022data}. They showed that the mean-squared error (MSE) of the reconstruction of the snapshots using an autoencoder (AE) for dimension reduction for the domain size of $L = 22$ exhibited an orders-of-magnitude drop when the dimension of the inertial manifold is reached. Furthermore, modeling the dynamics with a dense NN at this dimension either with a discrete-time map \cite{linot2020deep} or a system of ordinary differential equations (ODE) \cite{linot2022data} yielded excellent trajectory predictions and long-time statistics. The approach they introduced is referred to here as \textit{data-driven  manifold  dynamics} (DManD). For the KSE in larger domains, it was found that simply observing MSE vs.~autoencoder bottleneck dimension was not sufficient to determine the manifold dimension and exhaustive tests involving time evolution models vs.~dimension were necessary to estimate the manifold dimension. P\'erez De Jes\'us \& Graham extended this approach to two-dimensional Kolmogorov flow in a chaotic regime \cite{perezde2023data}. Here dimension reduction from 1024 dimensions to $\leq 10$ was achieved, with very good short-time tracking predictions, long-time statistics, as well as accurate predictions of bursting events. Linot \& Graham considered three-dimensional direct numerical simulations of turbulent Couette flow at $\text{Re}=400$ and found accurate data-driven dynamic models  with fewer than 20 degrees of freedom \cite{Linot.2023.10.1017/jfm.2023.720}. These models were able to capture characteristics of the flows such as streak breakdown and regeneration, short-time tracking, as well as Reynolds stresses and energy balance. They also computed unstable periodic orbits from the models with close resemblance to previously computed orbits from the full system. {Relatedly}, Zeng et al. \cite{Zeng.2024.10.1088/2632-2153/ad4ba5} exploited advances in autoencoder architecture \cite{jing2020implicit} to yield more precise estimates of $d_\IM$ for data from high-dimensional chaotic systems. The present work illustrates how symmetries of a flow system can be exploited in the DManD framework to yield highly efficient low-dimensional data-driven models for chaotic flows. \MDGrevise{In particular, incorporating symmetries both guarantees that the reduced order model automatically respects them and dramatically increases the effective density of data sampling.}

A fundamental notion in the topology of manifolds, which will be useful in exploiting symmetry, is that of charts and atlases \cite{lee2013smooth}. {Simply put, a chart is a region of a manifold whose points can be represented in a local Cartesian coordinate system of dimension $d_\IM$ and which overlaps with neighboring charts, while an atlas is a collection of charts that covers the manifold. \MDGrevise{(Strictly speaking, a chart is the region \emph{and} the local coordinate representation of each point in the region. We do not emphasize the latter aspect here because we will always have at hand explicit coordinates.)} This representation of a manifold has several advantages. First, for a manifold with dimension $d_\IM$, it may not be possible to find a \emph{global} coordinate representation in $d_\IM$ dimensions: Whitney's embedding theorem states that generic smooth maps for a smooth manifold of dimension $d_\mathcal{M}$ can be embedded into a Euclidean space of $2d_\mathcal{M}$.  Dividing a manifold into an atlas of charts enables \emph{minimal-dimensional} representations locally in $d_\IM$ dimensions.} Second, from the dynamical point of view, dynamics on different parts of a manifold may be very different and a single global representation of a manifold and the dynamics on it may not efficiently capture the dynamics, especially in the data-driven context. These two advantages of an atlas-of-charts representation have been exploited for data-driven modeling. {Floryan \& Graham developed a method to implement data-driven local representations for dynamical systems such as the quasiperiodic dynamics on a torus, a reaction-diffusion system, and the KSE to learn dynamics on invariant manifolds of minimal  dimension} \cite{floryan2021charts}. They refer to this method as \emph{Charts and Atlases for Nonlinear Data-Driven Dynamics on Manifolds} -- ``CANDyMan". Fox et al. then applied this to the MFE model, which displays highly intermittent behavior in the form of quasilaminarization and full relaminarization events, demonstrating more accurate time evolution predictions as global (``single chart") model \cite{fox2023predicting}. In the present work, we use the charts and atlases framework to exploit symmetries.

\newcommand{\boldy}{\boldsymbol{y}}
\newcommand{\boldq}{\boldsymbol{q}}
In \MDGrevise{partial differential equations (PDEs) such as }the Navier-Stokes equations (NSE), symmetries appear in the form of continuous and discrete operations such as translations, reflections, and rotations,  under which the equations are unchanged, and which map solutions to solutions. \MDGrevise{For example, for flow in a duct with a rectangular cross-section, the reflection of a solution to the problem across the symmetry planes of the cross section is also a solution to the problem. For flow in a circular pipe, the rotation of a solution by any angle around the central axis is another solution, and likewise for reflection across any plane cutting through that axis. For Kolmogorov flow, the example considered here, the symmetries arise from the nature of the forcing and the periodic boundary conditions, as detailed in Section \ref{sec:setup}. We emphasize that the approach described here is data-driven, relying only on knowledge the appropriate symmetries, not on the specific governing equations.} 

\begin{figure}
	\centering
	\includegraphics[width=0.9\linewidth]{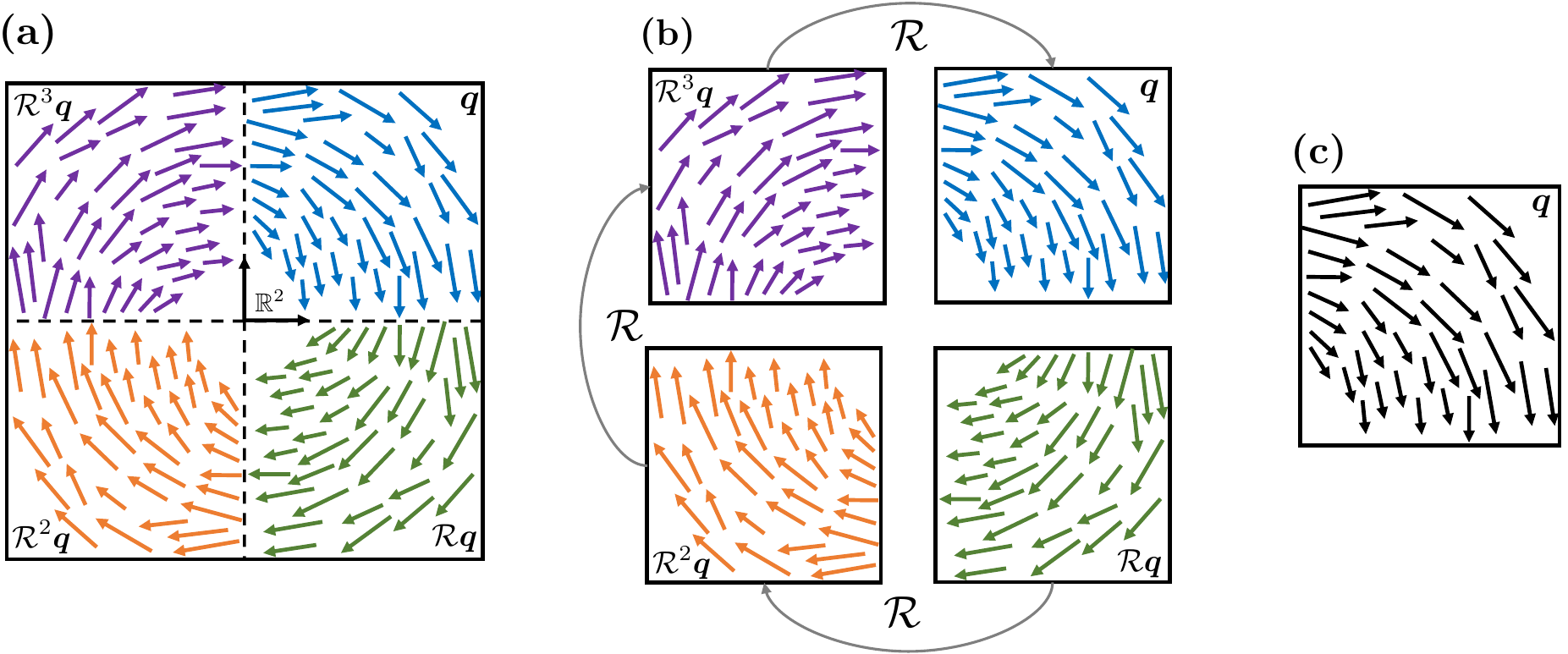}
	\caption{(a) Vector field $\boldsymbol{q}$ with discrete symmetry group $\mathcal{R}$ in state space \MDGrevise{ or on an invariant manifold within a state space}. (b) Points and vectors in each quadrant can be mapped to another quadrant by operating with $\mathcal{R}$. (c) Since the vector field at any point in the domain can be mapped into the first quadrant, this region of state space can be taken to be the fundamental domain.  Extending its boundaries to overlap with the other quadrants yields the fundamentl chart for this vector field.}
	\label{fig:State_sym}
\end{figure}

Symmetries in physical space  are reflected in state space as symmetries of the vector field, and these can generate natural charts of the system. In this work we leverage knowledge of the symmetries  to construct reduced-order models that automatically satisfy them. \MDGrevise{As an illustrative example, consider points $\boldy$ in state space $\reals^2$ (or equivalently, in a two-dimensional invariant manifold in a higher-dimensional state space) and the vector field $\boldsymbol{q}(\boldy)$ shown in Fig. \ref{fig:State_sym}\textcolor{blue}{a}. This field is equivariant with respect to the operation $\mathcal{R} \boldsymbol{q}$ that rotates the vector field by $\pi/2$ in the clockwise direction around the origin. That is, $\boldq(\mathcal{R}\boldy)=\mathcal{R}\boldq(\boldy)$. } Trajectories of the ODE $\dot{\by}=\bq$ will reflect this symmetry. Without exploiting symmetry, any model trained on trajectory data $\boldy(t)$ will need to represent the whole state space and even so will generally not obey the exact symmetries of the true system due to finite sampling effects.   We show a depiction of in Fig.~\ref{fig:State_sym}\textcolor{blue}{b} where we know that by applying $\mathcal{R}$ to the data we can map it to a different quadrant, and that by mapping points $\boldy$ and corresponding vectors $\boldq$ in quadrants 2, 3, and 4 with $\mathcal{R}$, $\mathcal{R}^2$, and $\mathcal{R}^3$, respectively, they all end up in quadrant 1. This is shown in Fig.~\ref{fig:State_sym}\textcolor{blue}{c}, where the vectors collapse on top of each other when mapped to the first quadrant. We will call this region the ``fundamental domain" for the state space (or an invariant manifold in the state space), and when expanded to overlap with its neighbors, the ``fundamental chart" \cite{Budanur2017}.  \MDGrevise{For each quadrant $j$ we can now define an ``indicator"  $\mathcal{I}=j-1$, such that a point $\boldy_j$ in quadrant $j=\mathcal{I}+1$ can be uniquely written in terms of a point $\boldy_1\equiv\boldy_F$ in the first quadrant -- i.e.~in the fundamental domain, with symmetries ``factored out" --  and the indicator $\mathcal{I}$, via the relation $\boldy_F=\mathcal{O}_\mathcal{I}\boldy_{\mathcal{I}+1}$, with $\mathcal{O}_\mathcal{I}=\mathcal{R}^\mathcal{I}$. Likewise for the corresponding vector field $\boldq$.  That is, the ``state-indicator'' pair $\left\{\boldy_F,\mathcal{I}\right\}$ uniquely determines each point in state space. Thus a trajectory $\boldy(t)$ that evolves under $\dot{\boldy}=\boldq$ can be written as $\left\{\boldy_F(t),\mathcal{I}(t)\right\}$. The present work generalizes this formalism. }

Some previous work has focused on factoring out symmetries of dynamical systems for dimension reduction in the data-driven context. However, symmetries have not been considered explicitly for data-driven reduced-order models (ROMs). Kneer et al. built symmetries into an AE architecture for dimension reduction and applied it to the Kolmogorov flow system \cite{kneer2021symmetry}. With the use of branches that receive the different discrete symmetries of the snapshots and spatial transformer networks that manipulate the continuous phase, they were able to map to a fundamental domain by selecting the branch that gives the smallest MSE and backpropagating through it. By doing this, the AE naturally selects a path that leads to the lowest MSE of reconstruction while considering the symmetries of the system. The purpose of this work was not to find the minimal dimension or learn ROMs. Budanur \& Cvitanovic formulated a symmetry-reducing scheme, previously applied to the Lorenz equations \cite{miranda1993proto}, to the KSE, where a polynomial transformation of the Fourier modes combined with the method-of-slices for factoring out spatial phase leads to a mapping in the fundamental domain \cite{Budanur2017}. Applying this method to more complicated systems, such as Kolmogorov flow, may be quite challenging. Symmetry reduction has also been successfully used for controlling the KSE with synthetic jets. Zeng \& Graham applied discrete symmetry operations to map the state to the fundamental domain, and used this as the input to a reinforcement learning control agent, the performance of which is dramatically improved by exploitation of symmetry\cite{zeng2021symmetry}.  


In the present work, we present a framework for learning DManD models in the fundamental chart, as in Fig. \ref{fig:State_sym}\textcolor{blue}{c}, where the dynamics of the flow occur, and apply it to chaotic dynamics of the NSE. {We refer to this method as ``symmetry charting'',  which results in (at least approximately) minimal-dimensional data-driven models that are equivariant to symmetry operations.} In addition the model only need to be learned in one chart because a combination of its symmetric versions will cover the manifold. \CEP{This approach is similar to the CANDyMan methodology, where now the charts are related by symmetry operations. 
}
{We first learn minimal-dimensional representations in the fundamental chart followed by an evolution equation for the dynamics on it.} To this end we use information about the symmetries of the system combined with NNs to learn the compressed representation and dynamics. 


We apply symmetry charting to {data from numerical simulations of} two-dimensional Kolmogorov flow. \MDGrevise{This flow is chosen because it is well-studied as a model of a flow with complex dynamics \cite{meshalkin1961investigation,chandler2013invariant,platt1991investigation,
armbruster1992phase,thess1992instabilities,inubushi2012covariant,
Farazmand2016,Page2020,Page2020,perezde2023data} while being relatively inexpensive to simulate.  Additionally, an experimental realization of this flow has been developed \cite{Suri.2020.10.1103/physrevlett.125.064501}.}  We focus on a Reynolds number regime where chaotic trajectories travel between unstable relative periodic orbits (RPOs) \cite{crowley2022turbulence} through bursting events \cite{armbruster1996symmetries} that shadow heteroclinic orbits connecting the RPOs. RPOs correspond to periodic orbits in a moving reference frame, such that in a fixed frame, the pattern at time $t + T$ is a phase-shifted replica of the pattern at time $t$. 
{To \CEP{perform symmetry charting} we map the data to a fundamental chart which is then used to train undercomplete AEs to find low-dimensional representations.} To this end, we use Implicit Rank Minimizing autoencoders with weight decay (IRMAE-WD) \cite{jing2020implicit,Zeng.2024.10.1088/2632-2153/ad4ba5}. {This architecture enables estimation of $d_\mathcal{M}$ without the need to sweep over dimensions and using MSE and statistics as indicators} \cite{linot2020deep,perezde2023data}. {We give an overview of IRMAE-WD in Section} \ref{sec:AE}. After learning a {low-dimensional} representation with IRMAE-WD we learn a time map with the use of the neural ODE (NODE) technique\cite{chen2018neural}, which we discuss in Section \ref{sec:time_evol}. We then show results in Section \ref{sec:Results} and finish with concluding remarks in Section \ref{sec:Conclusion}. Appendix \ref{sec:AppendixA} includes details of applying symmetry operations in Fourier space, and Appendix \ref{sec:channelAppendix} outlines how the symmetry charting method described here applies to 3D plane channel flow.

\section{Kolmogorov flow, symmetries, and projections}\label{sec:setup}

{The ``data set" we consider in applying symmetry charting consists of snapshots of the vorticity field from numerical solutions of the two-dimensional (2D) Navier-Stokes equations with Kolmogorov forcing:}
\begin{gather}
\frac{\partial \boldsymbol{u}}{\partial t}+\boldsymbol{u} \cdot \grad \boldsymbol{u}+\grad p=\frac{1}{\Rey} \nabla^{2} \boldsymbol{u}+\sin (n y) \hat{\boldsymbol{x}},  \\
\nabla \cdot \boldsymbol{u}=0,
\end{gather}
where flow is in the $x-y$ plane, $\boldsymbol{u}=[u,v]$ is the velocity vector, $p$ is the pressure, $n$ is the wavenumber of the forcing, and $\hat{\boldsymbol{x}}$ is the unit vector in the $x$ direction. Here $\Rey=\frac{\sqrt{\chi}}{v}\left(\frac{L_{y}}{2 \pi}\right)^{3 / 2}$ where $\chi$ is the dimensional forcing amplitude, $\nu$ is the kinematic viscosity, and $L_y$ is the size of the domain in the $y$ direction. We consider the square periodic domain $[-\pi,\pi] \times[-\pi, \pi]$. Vorticity is defined as $\omega = \hat{\boldsymbol{z}}\cdot\grad \times \boldsymbol{u}$, where $ \hat{\boldsymbol{z}}$ is the unit vector in the $z$ direction (orthogonal to the flow). \MDGrevise{Because of the symmetries of the forcing term $\sin(ny)\hat{\boldsymbol{x}}$, along with the rectangular periodic shape of the domain}, the equations are invariant under several symmetry operations \cite{chandler2013invariant}, namely a shift (in $y$)-reflect (in $x$) $\mathscr{S}$, a rotation $\mathscr{R}$ through $\pi$, and a continuous translation $\mathscr{T}_l$ over an arbitrary distance $l$ in $x$:
\begin{gather}
\mathscr{S}:[u, v, \omega](x, y) \rightarrow[-u, v,-\omega]\left(-x, y+\frac{\pi}{n}\right), \label{eq:SR}\\
\mathscr{R}:[u, v, \omega](x, y) \rightarrow[-u,-v, \omega](-x,-y), \label{eq:Rot}\\
\mathscr{T}_{l}:[u, v, \omega](x, y) \rightarrow[u, v, \omega](x+l, y).\label{eq:trans}
\end{gather}
For a forcing with $y$-wavenumber $n$, the state can be shift-reflected $2n-1$ times with the operation $\mathscr{S}$. Together with the rotation $\mathscr{R}$ there will be a total of $4n$ states that are related by symmetries. In the case of $n=2$ this results in eight regions of state space that are related to one another by these discrete symmetries.  In the chaotic regime that we consider, trajectories visit each of these regions. Relatedly, velocity fields for the Kolmogorov flow satisfy the \emph{equivarance} property associated with these actions: if $\bu$ is a solution at a given time, then so are $\mathscr{S}\bu$, $\mathscr{R}\bu$, and $\mathscr{T}_l\bu$. We elaborate on this property in Sec.~\ref{sec:prepros}.

The total kinetic energy for this system ($KE$), dissipation rate ($D$) and power input ($I$) are 
\begin{equation}
KE =\frac{1}{2}\left\langle\boldsymbol{u}^{2}\right\rangle_{V}, D=\frac{1}{\Rey}\left\langle|\nabla \boldsymbol{u}|^{2}\right\rangle_{V}, \quad I=\langle u \sin (n y)\rangle_{V}
\end{equation}
where $\left\langle \cdot \right\rangle_{V}$ corresponds to the average taken over the domain. For the case of $n=1$ the laminar state is linearly stable at all $\text{Re}$ \cite{iudovich1965example}. It is not until $n=2$ that the laminar state becomes unstable, with a critical value of $\Rey_c=n^{3/2}2^{1/4}$\cite{meshalkin1961investigation, green1974two, thess1992instabilities}. In what follows, we evolve the NSE numerically in the vorticity representation on a $[d_x \times d_y]=[32 \times 32]$ grid following the pseudo-spectral scheme given by Chandler \& Kerswell \cite{chandler2013invariant}, which is based on the code by Bartello \& Warn \cite{bartello1996self}. {We find that this resolution is sufficient at the conditions considered. We emphasize however, that the aim of the present work is not to use machine learning to solve the NSE, but rather to find a high-fidelity low-dimensional representation of the dynamics from a \emph{data set}, which in this case comes from the numerical solution described here.}

\begin{figure} 
	\centering
	\begin{subfigure}{.5\textwidth}
		\centering
		\includegraphics[width=1\linewidth]{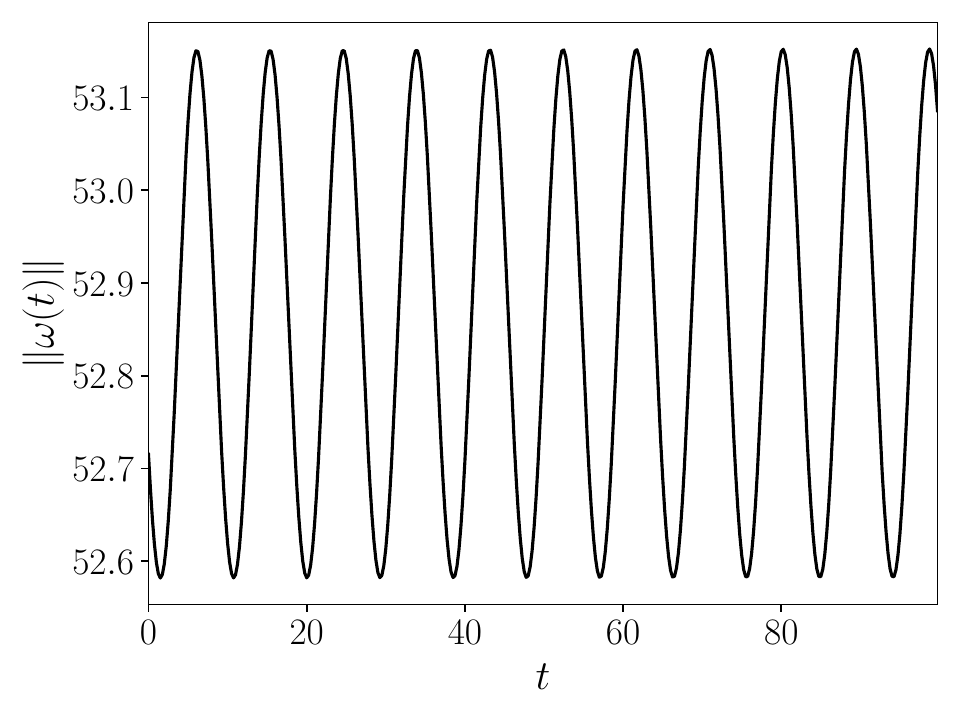}
		\caption{}
		\label{fig:ens_true_13d5}
	\end{subfigure}%
	\begin{subfigure}{.5\textwidth}
		\centering
		\includegraphics[width=1\linewidth]{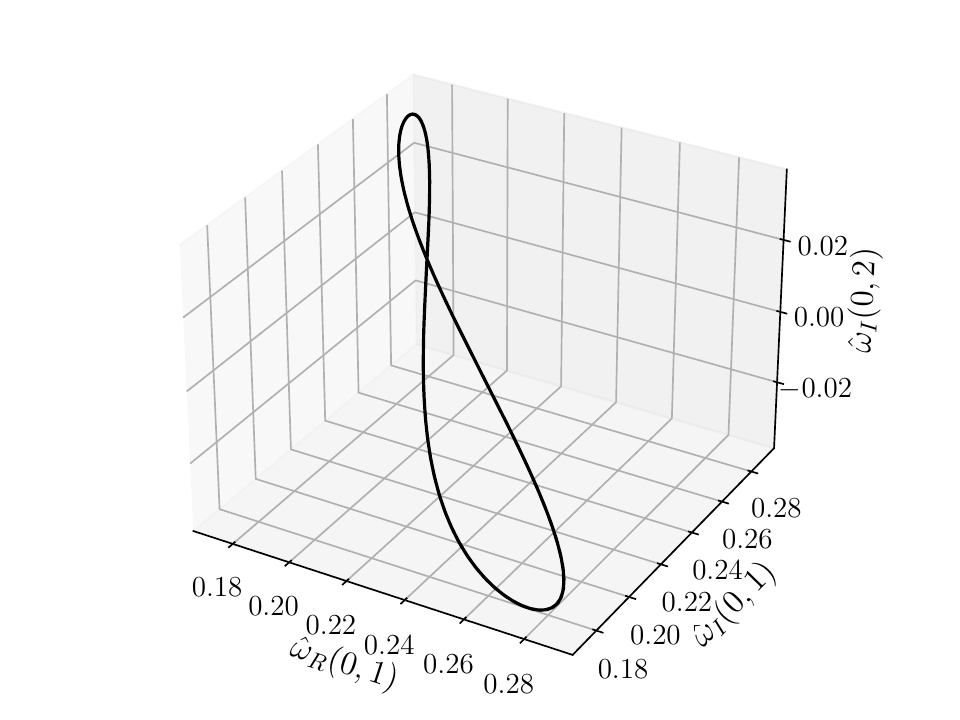}
		\caption{}
		\label{fig:3dproj_true_13d5}
	\end{subfigure}
	\caption{(a) Time evolution of $\| \omega (t) \| $ at $\Rey=13.5$. (b) State-space projection of the trajectory into the subspace $\hat{\omega}_R(0, 1)-\hat{\omega}_I(0, 1)-\hat{\omega}_I(0, 2)$ for $\Rey=13.5$. }
	\label{fig:trueRe13d5}
\end{figure}

\begin{figure} 
	\centering
	\begin{subfigure}{.5\textwidth}
		\centering
		\includegraphics[width=1\linewidth]{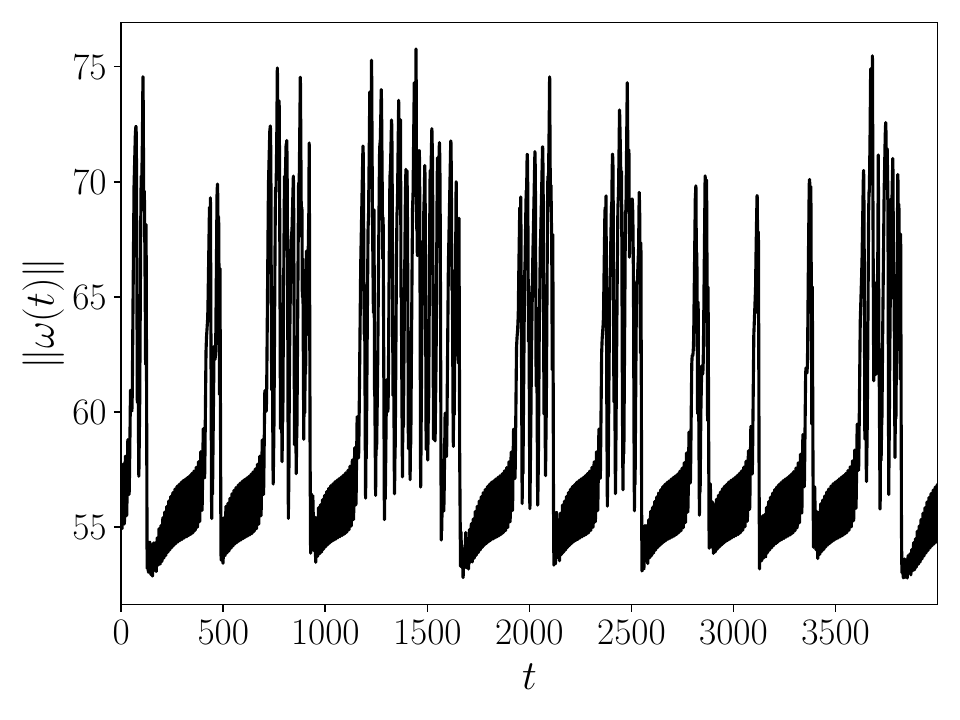}
		\caption{}
		\label{fig:ens_true_14d4}
	\end{subfigure}%
	\begin{subfigure}{.5\textwidth}
		\centering
		\includegraphics[width=1\linewidth]{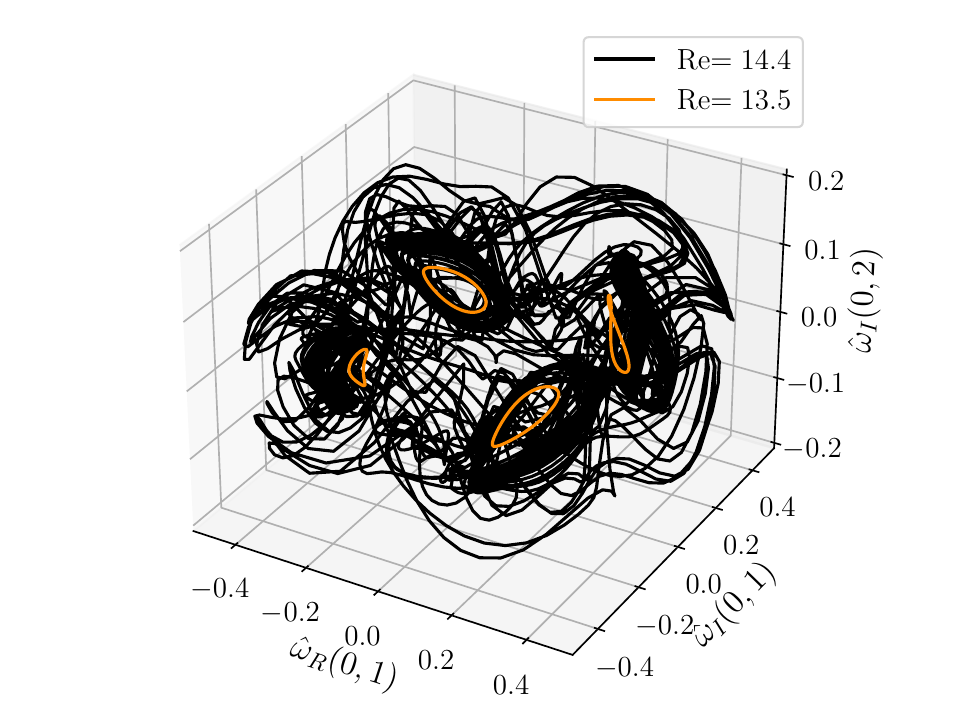}
		\caption{}
		\label{fig:3dproj_true_14d4}
	\end{subfigure}
	\caption{(a) Time evolution of $\| \omega (t) \| $ at $\Rey=14.4$. (b) State-space projection of the trajectory into the subspace $\hat{\omega}_R(0, 1)-\hat{\omega}_I(0, 1)-\hat{\omega}_I(0, 2)$ for $\Rey=13.5$ and $\Rey=14.4$. }
	\label{fig:trueRe14d4}
\end{figure}

In Fig.\ \ref{fig:trueRe13d5} we show the time-series evolution of $\| \omega (t) \| $, where $\| \cdot \|$ is the $l^2$-norm, for an RPO obtained at $\text{Re}=13.5$, $n=2$, as well as the state-space projection of the trajectory into the subspace $\hat{\omega}_R(0, 1)-\hat{\omega}_I(0, 1)-\hat{\omega}_I(0, 2)$ where $\hat{\omega} (k_x,k_y,t) =\mathcal{F}\{\omega(x,y,t)\} = \hat{\omega}_R(k_x, k_y,t)+i\hat{\omega}_I(k_x, k_y,t)$ is the discrete Fourier transform in $x$ and $y$, and subscripts $R$ and $I$ correspond to real and imaginary parts. We also consider chaotic Kolmogorov flow with $\text{Re}=14.4$, $n=2$. Fig. \ref{fig:trueRe14d4} shows the time-series evolution of $\| \omega (t) \| $ as well as the state-space projection of the trajectory into the subspace $\hat{\omega}_R(0, 1)-\hat{\omega}_I(0, 1)-\hat{\omega}_I(0, 2)$ for this chaotic regime. Due to the discrete symmetries of the system, there are several RPOs \cite{armbruster1996symmetries}. The dynamics are characterized by quiescent intervals where the trajectories approach the RPOs (which are now unstable), punctuated by fast excursions between the RPOs, which are indicated by the intermittent increases of the $\| \omega (t) \| $ in Fig. \ref{fig:ens_true_14d4}. Under the projection shown in Fig. \ref{fig:3dproj_true_14d4}, we see four RPOs, which initially seems surprising as there are eight discrete symmetries and a continuous symmetry. However, the continuous symmetry is removed under this projection (it will only appear for $k_x \neq 0$), and the discrete symmetry operations of $\mathscr{R}$ and $\mathscr{S}$ will flip the signs of $\hat{\omega}_R(0, 1)$, $\hat{\omega}_I(0, 1)$, and $\hat{\omega}_I(0, 2)$ such that portions of the different RPOs are covered. Specifics of the sign changes will be made clear in Section \ref{sec:prepros}. 



\section{Data-driven dimension reduction and dynamic modeling} \label{sec:AEs}

In the following subsections, we describe the steps involved in symmetry charting: 1) \MDGrevise{developing criteria to define a}  fundamental \MDGrevise{chart in the the original (high-dimensional) state space and what symmetry operations are need to map any point in state space to that chart}, 2) {finding a minimal-dimensional coordinate representation, \MDGrevise{i.e.~the fundamental chart for the invariant manifold}, and 3) evolving \ALrevise{both} the minimal-dimensional state and \ALrevise{the} symmetry indicators forward through time.} The parts of this procedure separating it from other data-driven reduced-order models are how we map to the fundamental domain and how we evolve the symmetry indicators through time. \ALrevise{Our approach guarantees that the reduced-dimensional model maintains  the equivariance properties of the original dynamical system.} 


\subsection{Map to fundamental domain}\label{sec:prepros}

\MDGrevise{In the example shown in Fig.~\ref{fig:State_sym}, only a simple, discrete symmetry was present, and it was straightforward to identify an indicator value $\mathcal{I}$ corresponding with each symmetry operation that would map a general state point $\boldy$ to the fundamental domain. This subsection illustrates how to do so for the symmetries of the Kolmogorov flow. To illustrate the generality of the approach, in Appendix \ref{sec:channelAppendix}, we outline the same process for plane channel flow. As noted above, for the Kolmogorov flow we consider, there are eight distinct discrete symmetry operations, so we define an indicator $\mathcal{I}\in\left\{0,1,2,\ldots, 7\right\}$ as well as the continuous translation in $x$, which we treat as a spatial phase shift $\phi_x$. We describe here how to take an arbitrary vorticity field $\omega(x,y)$ and rewrite it uniquely as the set $\left\{\omega_F(x,y),\phi_x,\mathcal{I}\right\}$, where $\omega_F(x,y)$ is a state point in a chosen fundamental domain, i.e.~with the discrete and continuous symmetries factored out. Now a trajectory $\omega(x,y,t)$ can be represented as $\left\{\omega_F(x,y,t),\phi_x(t),\mathcal{I}(t)\right\}$. }

\MDGrevise{The criteria that determine $\phi_x$ and the various indicator indices $\mathcal{I}$ are most conveniently represented in Fourier space.} 
With $x$- and $y$-wavenumbers $k_x$ and $k_y$, respectively, operations \eqref{eq:SR}-\eqref{eq:trans} applied to the vorticity field at a particular time instant become 
\begin{equation}  \label{eq:SR_fourier_c}
	\mathscr{S}:\hat{\omega}(k_x, k_y) \rightarrow -\hat{\omega}(-k_x, k_y)e^{i k_y\pi/n},
\end{equation}
\begin{equation} \label{eq:R_fourier_c}
	\mathscr{R}:\hat{\omega}(k_x, k_y) \rightarrow \bar{\hat{\omega}}(k_x, k_y),
\end{equation}
\begin{equation} \label{eq:shift_fourier_c}
	\mathscr{T}_l:\hat{\omega}(k_x, k_y) \rightarrow \hat{\omega}(k_x, k_y)e^{i k_x l},
\end{equation}
where $\hat{\cdot}$ denotes Fourier transform $\mathcal{F}$ and $\bar{\cdot}$ denotes complex conjugate. \MDGrevise{Moreover, since the vorticity field is real, $\hat{\omega}(k_x,-k_y)=\bar{\hat{\omega}}(-k_x, k_y)$ and $\hat{\omega}(-k_x,-k_y)=\bar{\hat{\omega}}(k_x, k_y)$.}

To compute the phase $\phi_x$ for a vorticity field,  we use the ``First Fourier mode method-of-slices'' \cite{budanur2015periodic, budanur2015reduction}. This method involves computing the spatial phase $\phi_x$ of the Fourier mode $k_x=1,k_y=0$: \MDGrevise{using standard trigonometric identities, this mode can be written as $2||\hat{\omega}(1,0)||\cos(x+\phi_x)$, with }$\phi_x(t)=\operatorname{atan} 2\left\{\hat{\omega}_I(1, 0), \hat{\omega}_R(1, 0)\right\}$, where subscripts $R$ and $I$ indicate real and imaginary part. Then, \MDGrevise{to factor out the translation}, we shift the vorticity snapshots such that this mode is a pure cosine: $\omega_{\phi_x}(x,y,t)=\mathcal{F}^{-1}\left\{\mathcal{F}\{\omega(x,y,t)\} e^{-i k_x \phi_x(t)}\right\}$ \MDGrevise{The translation of the solution in the $x$-direction is captured by the time-dependence of $\phi_x$.}

\MDGrevise{Now we turn to the discrete symmetries. As noted above, for a given state $\omega(x,y)$ there are seven other states related by the symmetry operations. Our aim is thus to identify eight regions of state space that are related to one another by these operations. We will define these regions using simple sign criteria based on how the dominant (low wavenumber) Fourier modes transform under $\mathscr{S}$ and $\mathscr{R}$. We will need three such criteria to identify the eight regions in state space.} \ALrevise{To simplify the analysis, we use coefficients with $k_x=0$ because these values are invariant to translations in $x$ and thus unchanged when the translation symmetry is factored out.}

\MDGrevise{The actions of $\mathscr{S}$ and $\mathscr{R}$ on $\hat{\omega}(0,1)$ and $\hat{\omega}(0,2)$ are as follows:}
\begin{equation} 
	\mathscr{S}:\hat{\omega}_R(0, 1)+i\hat{\omega}_I(0, 1) \rightarrow \hat{\omega}_I(0, 1)-i\hat{\omega}_R(0, 1),
\end{equation}
\begin{equation} 
	\mathscr{S}:\hat{\omega}_R(0, 2)+i\hat{\omega}_I(0, 2) \rightarrow \hat{\omega}_R(0, 2)+i\hat{\omega}_I(0, 2),
\end{equation}
\begin{equation}
	\mathscr{R}:\hat{\omega}_R(0, 1)+i\hat{\omega}_I(0, 1) \rightarrow \hat{\omega}_R(0, 1)-i\hat{\omega}_I(0, 1),
\end{equation}
\begin{equation}
	\mathscr{R}:\hat{\omega}_R(0, 2)+i\hat{\omega}_I(0, 2) \rightarrow \hat{\omega}_R(0, 2)-i\hat{\omega}_I(0, 2).
\end{equation}
\ALrevise{Observe that the shift-reflect operator $\mathscr{S}$ modifies the $(k_x=0,k_y=1)$, but not the $(k_x=0,k_y=2)$ coefficients. I.e.~$\mathscr{S}$ maps between different quadrants of the $\hat{\omega}_R(0, 1)- \hat{\omega}_I(0, 1)$ plane: $\mathscr{S}^3, \mathscr{S}^2$, and $\mathscr{S}$ map from quadrants 2, 3, and 4, of this plane to the first quadrant, respectively. The rotation operator $\mathscr{R}$  modifies both the  $(k_x=0,k_y=1)$ and $(k_x=0,k_y=2)$ coefficients by flipping the sign of the imaginary component. 
We now can take the fundamental domain to be the set of points whose projection into the $(\hat{\omega}_R(0, 1)$, $\hat{\omega}_I(0, 1)$, $\hat{\omega}_I(0, 2))$ lies in the $(+,+,+)$ octant, and use the above observations to determine what operation maps a point whose projection lies in any other octant into the first.
Table \ref{table_I} shows these discrete operations, identifying each octant by the signs of the Fourier coefficients.  For example, for a data point in the $(-,-,-)$ octant, the operation $\mathscr{R}\mathscr{S}^3$ maps it into the fundamental domain.
 \MDGrevise{The entries for $\mathscr{R},\mathscr{S},\mathscr{S}^2$ and $\mathscr{S}^3$ follow directly from the discussion above, with the remaining entries obtained by inspection. 
} }

\MDGrevise{In summary, for any vorticity field, the signs of $\hat{\omega}_R(0, 1)$, $\hat{\omega}_I(0, 1)$, $\hat{\omega}_I(0, 2)$ specify an octant, and one of the eight symmetry operations will bring the field into the first $(+,+,+)$ octant. } 
\MDGrevise{Now we can fully characterize the fundamental domain as the region in state space where $\phi_x=0$ and $\mathcal{I}=0$. For any vorticity field $\omega(x,y,t)$ we can determine and factor out $\phi_x(t)$ and $\mathcal{I}(t)$ using the above criteria to give a point $\omega_F(x,y,t)$.} 

\begin{table}[t]
\caption{Indicators $\mathcal{I}$, corresponding operations required to map to $\mathcal{I}=0$, the fundamental domain, and signs of the relevant Fourier amplitudes. } 
$$
\begin{array}{ccccc}\hline \hline \mathcal{I} & \text {\; \; \; \; Discrete operation } & \sgn(\hat{\omega}_R(0,1)) & \sgn(\hat{\omega}_I(0,1)) & \sgn(\hat{\omega}_I(0,2))\\ \hline  0 & \text{Identity} & + & + & +  \\ 
1 & \mathscr{R} & + & \CEP{-} & \CEP{-} \\  
2 & \mathscr{S} & - & + & + \\  
3 & \mathscr{R}\mathscr{S}^3 & - & \CEP{-} & - \\  
4 & \mathscr{R}\mathscr{S}^2 & - & \CEP{+} & - \\  
5 & \mathscr{R}\mathscr{S} & + & \CEP{+} & - \\  
6 & \mathscr{S}^2 & - & - & + \\  
7 & \mathscr{S}^3 & + & - & + \\ \hline \hline\end{array}
$$
\label{table_I}
\end{table}

\begin{figure}
	\centering
	\includegraphics[width=0.7\linewidth]{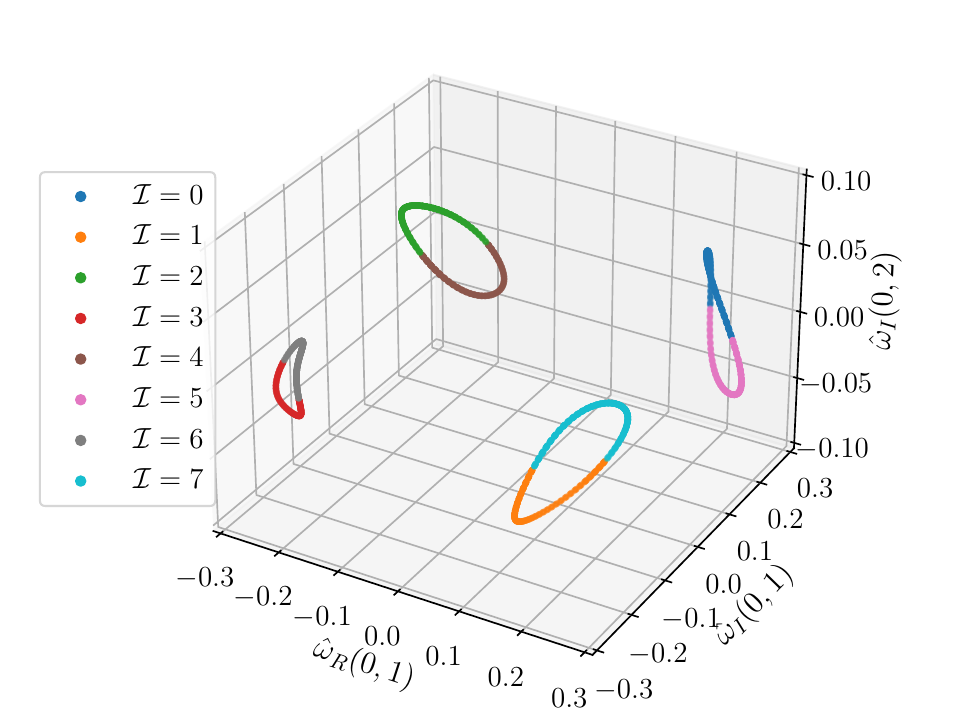}
	\caption{State-space projection of the different RPOs at $\Rey=13.5$ into the subspace $( \hat{\omega}_R(0, 1), \hat{\omega}_I(0, 1),\hat{\omega}_I(0, 2))$. \ALrevise{The colors indicate the discrete symmetry operations (Table \ref{table_I}) required to map the data to the fundamental domain.}
    }
	\label{Re13d5_allsym}
\end{figure}
\begin{figure}
	\centering
	\begin{subfigure}{.5\textwidth}
		\centering
		\includegraphics[width=1\linewidth]{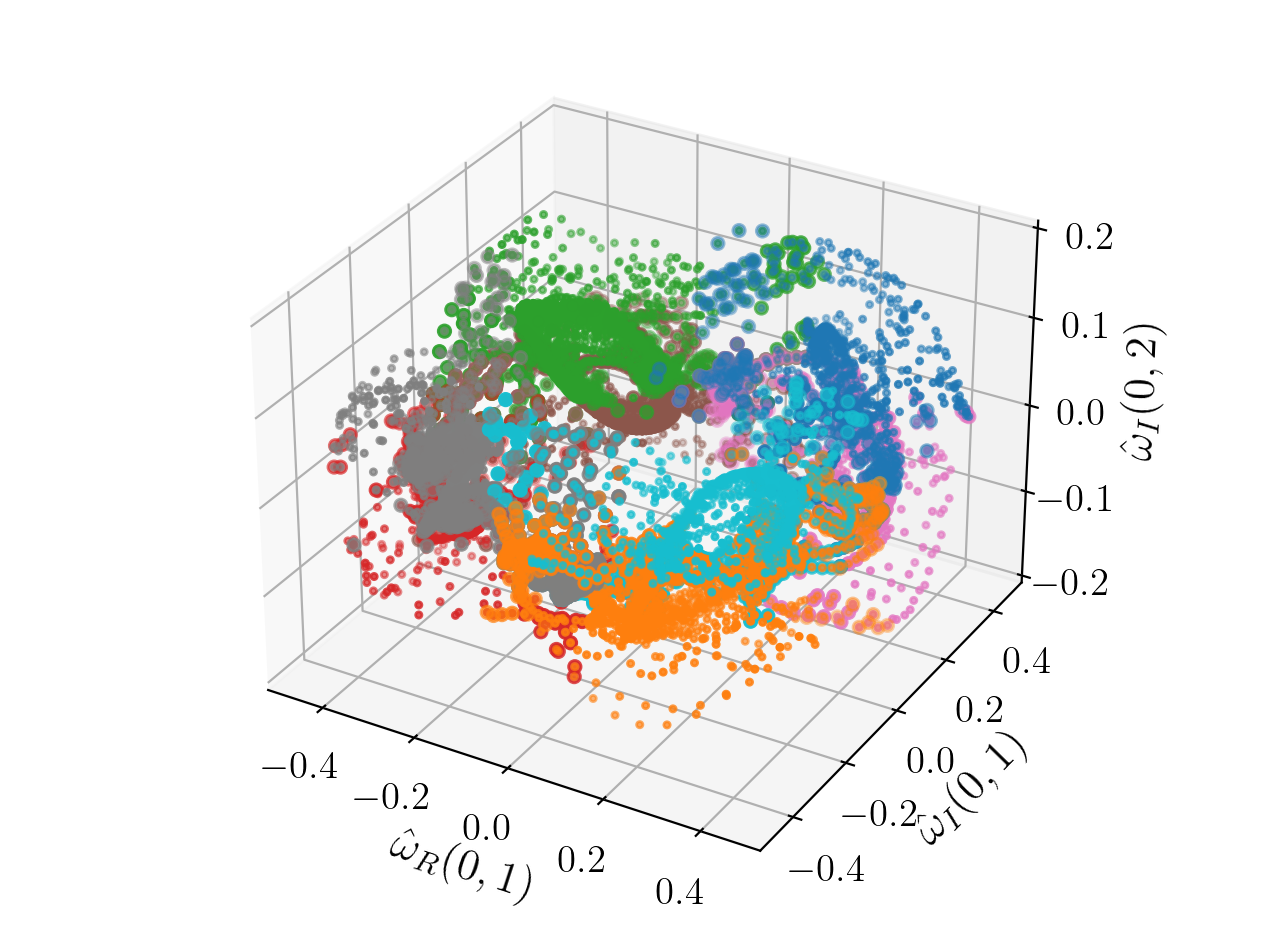}
		\caption{}
		\label{chart_blowupa}
	\end{subfigure}%
        \begin{subfigure}{.5\textwidth}
		\centering
		\includegraphics[width=1\linewidth]{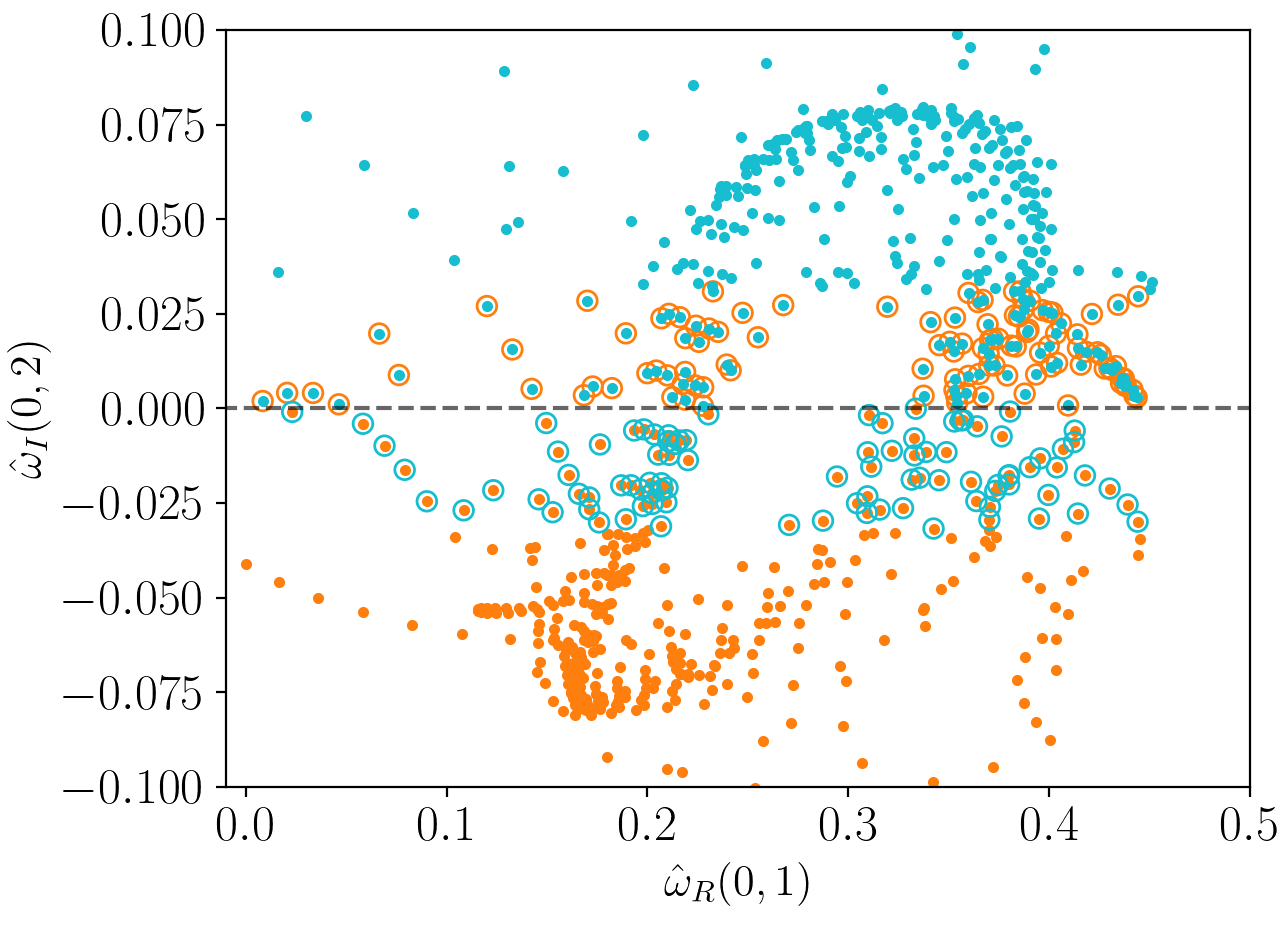}
		\caption{}
		\label{chart_blowupb}
	\end{subfigure}%
	\caption{ (a) Projection of a chaotic trajectory at $\Rey=14.4$ into the subspace $(\hat{\omega}_R(0, 1),\hat{\omega}_I(0, 1),\hat{\omega}_I(0, 2))$.  \ALrevise{The colors indicate the discrete symmetry operations (Table \ref{table_I}) required to map the data to the fundamental domain.} (b) Zoom into the region near the RPO corresponding to $\ind=1,7$. \MDGrevise{Colors of solid markers correspond to which octant they occupy. Points surrounded by open circles are interior points for one chart and exterior points to a neighboring one. E.g.~ the blue solid markers within an open orange marker are in the interior region of chart 7 and exterior region of chart 1 and vice versa.} }
	\label{chart_blowup}
\end{figure}

\MDGrevise{Fig. \ref{Re13d5_allsym} shows the state-space projection of four symmetry-related relative periodic orbits (RPOs) at $\Rey=13.5$ into the subspace $(\hat{\omega}_R(0, 1),\hat{\omega}_I(0, 1),\hat{\omega}_I(0, 2))$,  where the different colors give the indicator for the octant in which that data point lies. } Notice that each section of the different RPOs can be uniquely identified by $\mathcal{I}$. Fig. \ref{chart_blowup}(a) shows the same, for  a chaotic trajectory at $\text{Re}=14.4$. 



\MDGrevise{
As we see in Figures \ref{Re13d5_allsym} and \ref{chart_blowup}(a), a trajectory of the dynamics may pass through several or even all of these octants. To handle the transitions between these regions, we }
use the concept of charts and atlases that is fundamental to the study of manifolds \cite{lee2013smooth}, in a way that is closely related to that of Floryan \& Graham \citep{floryan2021charts}. An atlas for a manifold is a collection of regions of the manifold, each of which must overlap with its neighbors, that are called charts. In each chart, a local coordinate representation can be found, and for each pair of overlapping charts, a transition map takes coordinates in one chart to those in the neighboring one -- with this coordinate transformation there are no discontinuities. \MDGrevise{Note that one can define an atlas of charts on both the ambient state space $\reals^N$ and on the invariant manifold $\IM$ of the dynamics. (The phase variable can be taken to be an additional coordinate for either the ambient state space or the invariant manifold.) The fundamental chart for $\IM$ is just the portion of $\IM$ in the fundamental chart of the ambient space, and it is related to the other charts via the indicators $\mathcal{I}$.   In this section we focus on the ambient space. }

\MDGrevise{To construct the fundamental chart, we expand the fundamental domain so that it overlaps with neighboring regions of state space.}
\MDGrevise{That is,} the fundamental chart is simply the fundamental domain, comprised of ``interior points"\MDGrevise{, whose projection into the $(\hat{\omega}_R(0, 1),\hat{\omega}_I(0, 1),\hat{\omega}_I(0, 2))$ subspace lie in the first octant,} plus an overlap region  
 \MDGrevise{ containing ``exterior points" whose projections lie just outside the first octant. Specifically, for a given data set, we find the point whose projection into the $(\hat{\omega}_R(0, 1),\hat{\omega}_I(0, 1),\hat{\omega}_I(0, 2))$ subspace has largest magnitude, i.e.~that maximizes $\hat{\omega}_R(0, 1)^2+\hat{\omega}_I(0, 1)^2+\hat{\omega}_I(0, 2)^2$. Denoting the coordinates of this projected point as $a,b,c$, we then define a box in the first octant with one corner at the origin and the far corner at $(\abs{a},\abs{b},\abs{c})$. That will be the fundamental domain. To make the fundamental chart, we simply expand that box in all directions by $20\%$ so it overlaps into all seven other octants. Any point whose projection lies in the volume between the original (first octant) box and this expanded box is an exterior point. We varied the percent increase to $5\%$, $10\%$, and $30\%$ and observed no change in the results.} 
\MDGrevise{Expanding the other seven octants in the same manner generates eight overlapping charts.}
We choose a coordinate representation for all charts that corresponds to that for the fundamental chart, and 
 for a trajectory that leaves the fundamental domain with e.g. $\ind=4$ and reenters with $\ind=6$, the transition map is the function that moves the point in the fundamental chart from the ``exit" of the $\ind=4$ chart to the ``entrance" of the $\ind=6$ chart. 

\MDGrevise{ In Fig.\ \ref{chart_blowup} we show the state space projection of a chaotic trajectory at $\Rey=14.4.$ Points surrounded by open circles are interior points for one chart and exterior points to a neighboring one. E.g.~the blue solid markers within an open orange marker are in the interior region of chart 7 and exterior region of chart 1 and vice versa.}

\begin{figure} 
	\centering
	
	\begin{subfigure}{.5\textwidth}
		\centering
		\includegraphics[width=1\linewidth]{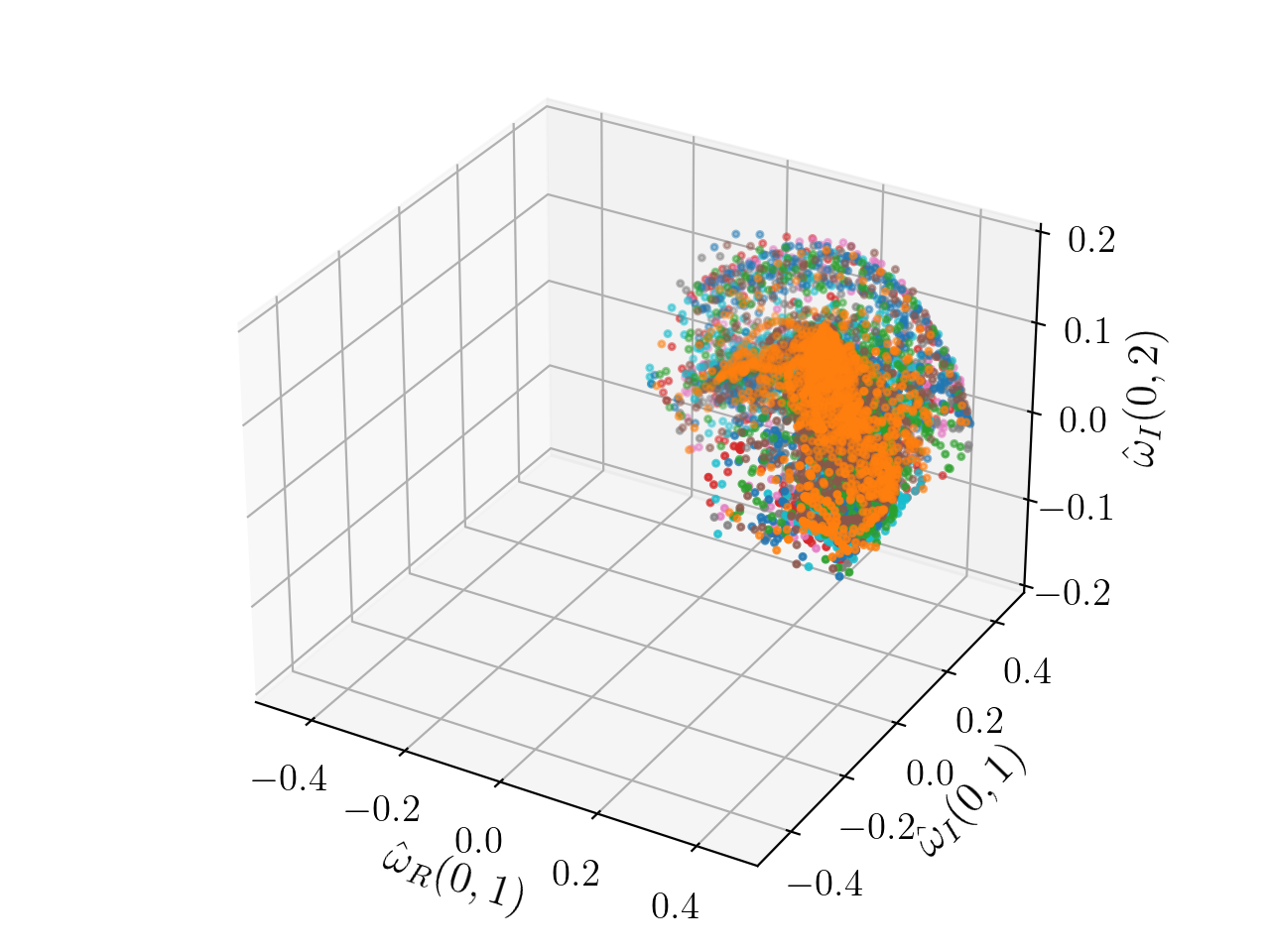} 
		\caption{}
		\label{fig:3dproj_14d4_colored_ppp}
	\end{subfigure}
        \begin{subfigure}{.5\textwidth}
		\centering
		\includegraphics[width=1\linewidth]{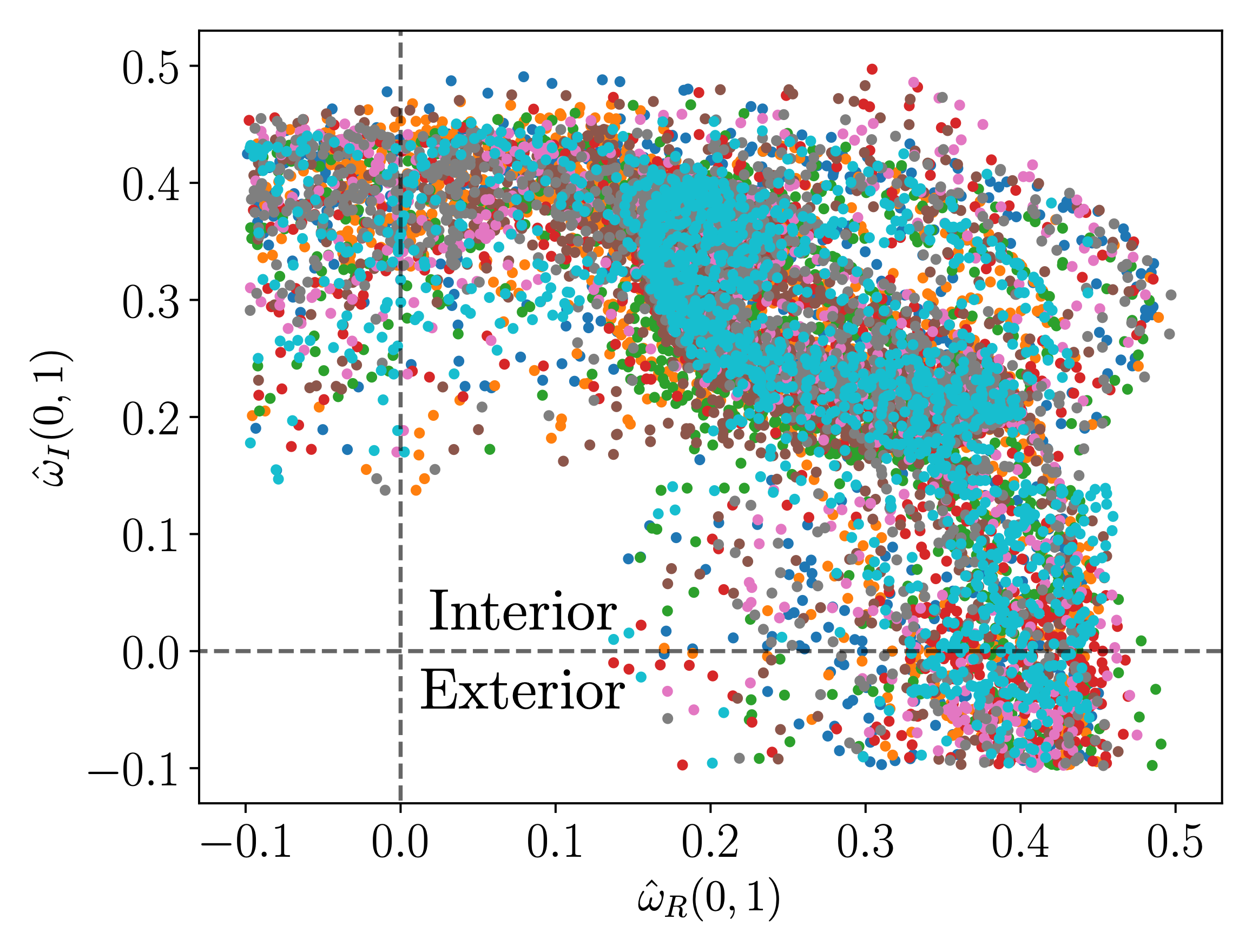}
		\caption{}
		\label{fig:2dproj_14d4_colored_re01im01}
	\end{subfigure}%
         \begin{subfigure}{.5\textwidth}
		\centering
		\includegraphics[width=1\linewidth]{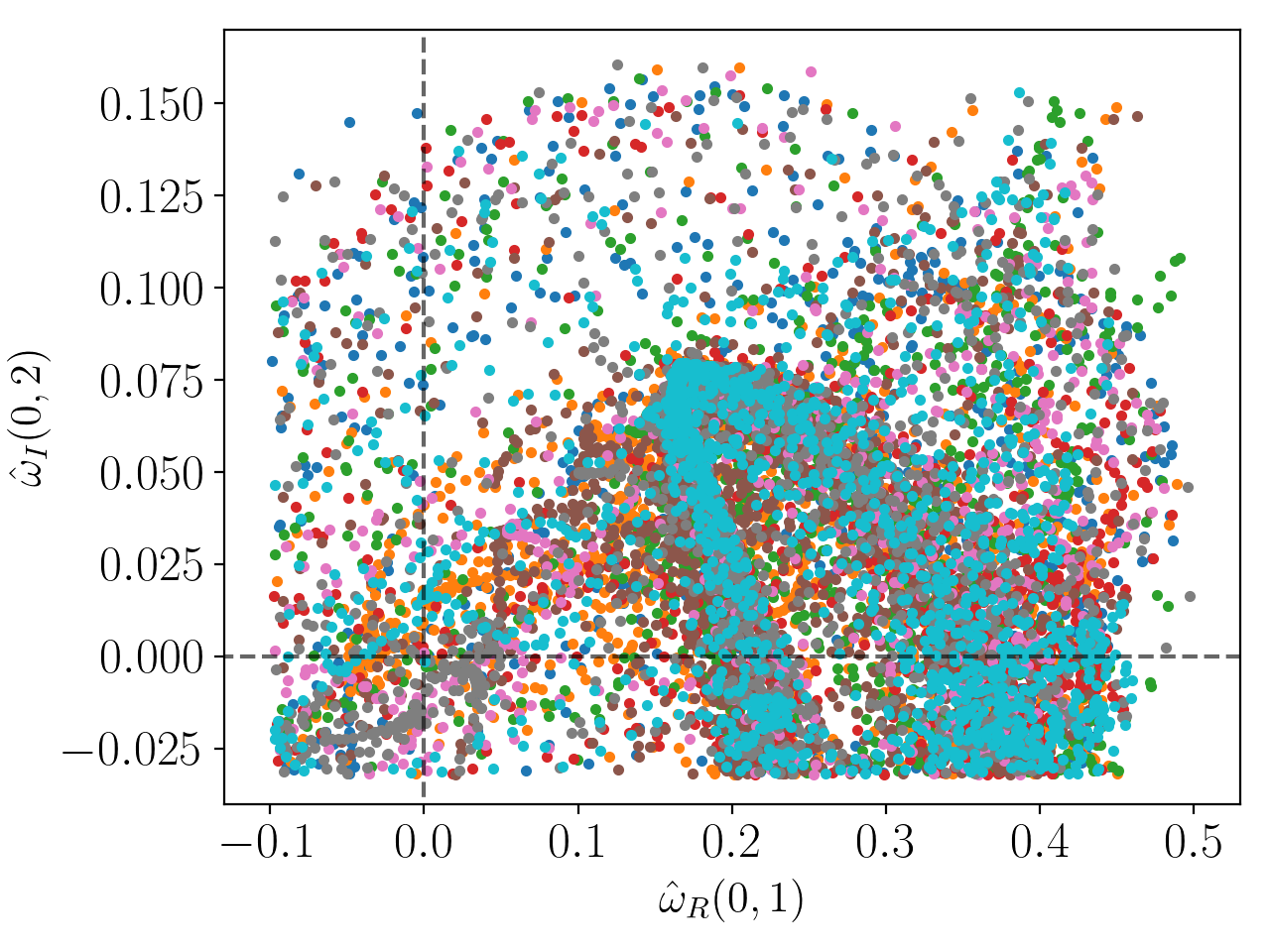}
		\caption{}
		\label{fig:2dproj_14d4_colored_re01im02}
	\end{subfigure}%
	\caption{ (a) Snapshots mapped to the fundamental chart  ($\hat{\omega}_R(0, 1)>0$, $\hat{\omega}_I(0, 1)>0$, $\hat{\omega}_I(0, 2)>0$) \MDGrevise{in the ambient space}. (b) and (c) are the state-space projection of the trajectory into the planes $\hat{\omega}_R(0, 1)-\hat{\omega}_I(0, 1)$ and  $\hat{\omega}_R(0, 1)-\hat{\omega}_I(0, 2)$, respectively. The dotted lines give the boundary between interior and exterior points.}
	\label{fig:trueRe14d4_colored}
\end{figure}

Now that we have identified these exterior points we can map the entire data set to the fundamental chart. To do this, we apply the appropriate discrete operations to the interior and exterior points of each chart to map the state such that the interior points fall in the first octant $\mathcal{I}=0$ ($\hat{\omega}_R(0, 1)>0$, $\hat{\omega}_I(0, 1)>0$, $\hat{\omega}_I(0, 2)>0$). For the chaotic flow at $\Rey=14.4$, Fig. \ref{fig:3dproj_14d4_colored_ppp} shows an entire trajectory mapped into the fundamental chart, with the colors indicating the indicator values for each point.  Figures \ref{fig:2dproj_14d4_colored_re01im01} and \ref{fig:2dproj_14d4_colored_re01im02} show projections in the planes $\hat{\omega}_R(0, 1)-\hat{\omega}_I(0, 1)$ and $\hat{\omega}_R(0, 1)-\hat{\omega}_I(0, 2)$, respectively, of the mapped data. The dashed lines correspond to the boundaries between the different octants. 
Observe that mapping the data to the fundamental chart leads to a much higher density of data points and thus to better sampling than if we considered the whole state space without accounting for the symmetries. \MDGrevise{From here on, we refer to data has been phase-aligned and mapped into the fundamental chart as $\tilde{\omega}$.\textbf{}}


\subsection{Finding a manifold coordinate representation with IRMAE-WD}\label{sec:AE}

In the previous section we demonstrated how to obtain the fundamental chart \MDGrevise{in the ambient space}. \MDGrevise{Now we turn to representing the lower-dimensional invariant manifold $\IM$, noting that we do not need to represent the entire manifold but only the portion that is in the fundamental chart of the ambient space, i.e.~the region illustrated in Fig.~\ref{fig:trueRe14d4_colored} . This will be the fundamental chart for $\IM$, with the other charts related to it via the symmetry operations.  Mapping all the data in the ambient space into the fundamental chart thus represents an eightfold augmentation of the data set that is available to determine the coordinate representation for the fundamental chart on $\IM$ (i.e.~in the autoencoder latent space).  }


{To find a low-dimensional representation of the fundamental chart on $\IM$} we use a variant of a common machine-learning architecture known as an undercomplete autoencoder (AE), whose purpose is to learn a reduced representation of the state such that the reconstruction error with respect to the true data is minimized. We flatten the vorticity field $\tilde{\omega}$ such that $N=32 \times 32 = 1024$.  The AE consists of an encoder and a decoder. The encoder maps from the full space $\tilde{\omega} \in \mathbb{R}^{N}$ to the lower-dimensional latent space $h(t) \in \mathbb{R}^{d_h}$, where ideally $d_h=d_\IM\ll N$, and the decoder maps back to the full space $\tilde{\omega}_r$. \MDGrevise{The vector $h(t)$ is a set of coordinates for representing the fundamental chart on $\IM$.}

{Previous works have focused on tracking MSE and dynamic model performance as $d_h$ varies to find good low-dimensional representations }\cite{linot2020deep,Linot.2023.10.1017/jfm.2023.720,perezde2023data}.  This is a tedious trial-and-error process that in general does not yield a precise estimate of $d_\IM$. Other works have learned compressed representations of flow problems \cite{Page2020, doan2021auto, nakamura2021convolutional}. However, performance over a systematic range of $d_h$ is not examined in these cases. A more systematic alternative would be highly desirable. In recent work, Zeng et al. \cite{Zeng.2024.10.1088/2632-2153/ad4ba5} showed that a straightforward variation on a standard autoencoder can yield robust  and precise estimates of $d_\IM$ for a range of systems, as well as an orthogonal manifold coordinate system. The architecture they study is called  an Implicit Rank Minimizing Autoencoder with weigh decay (IRMAE-WD), and  involves inserting a series of linear layers between the encoder and decoder and adding an $L_2$ regularization on the neural network weights in the loss. The effect of these additions is an AE for which the standard gradient descent algorithm for learning NN weights drives the rank of the covariance of the data in the latent representation to a minimum while maintaining representational capability. Applying this to the KSE and other systems resulted in the rank being equal to the dimension of the inertial manifold $d_\mathcal{M}$. IRMAE was originally presented by Jing et al. \cite{jing2020implicit} to learn low-rank representations for image-based classification and generative problems.

\begin{figure}
	\centering
	\includegraphics[width=1\linewidth]{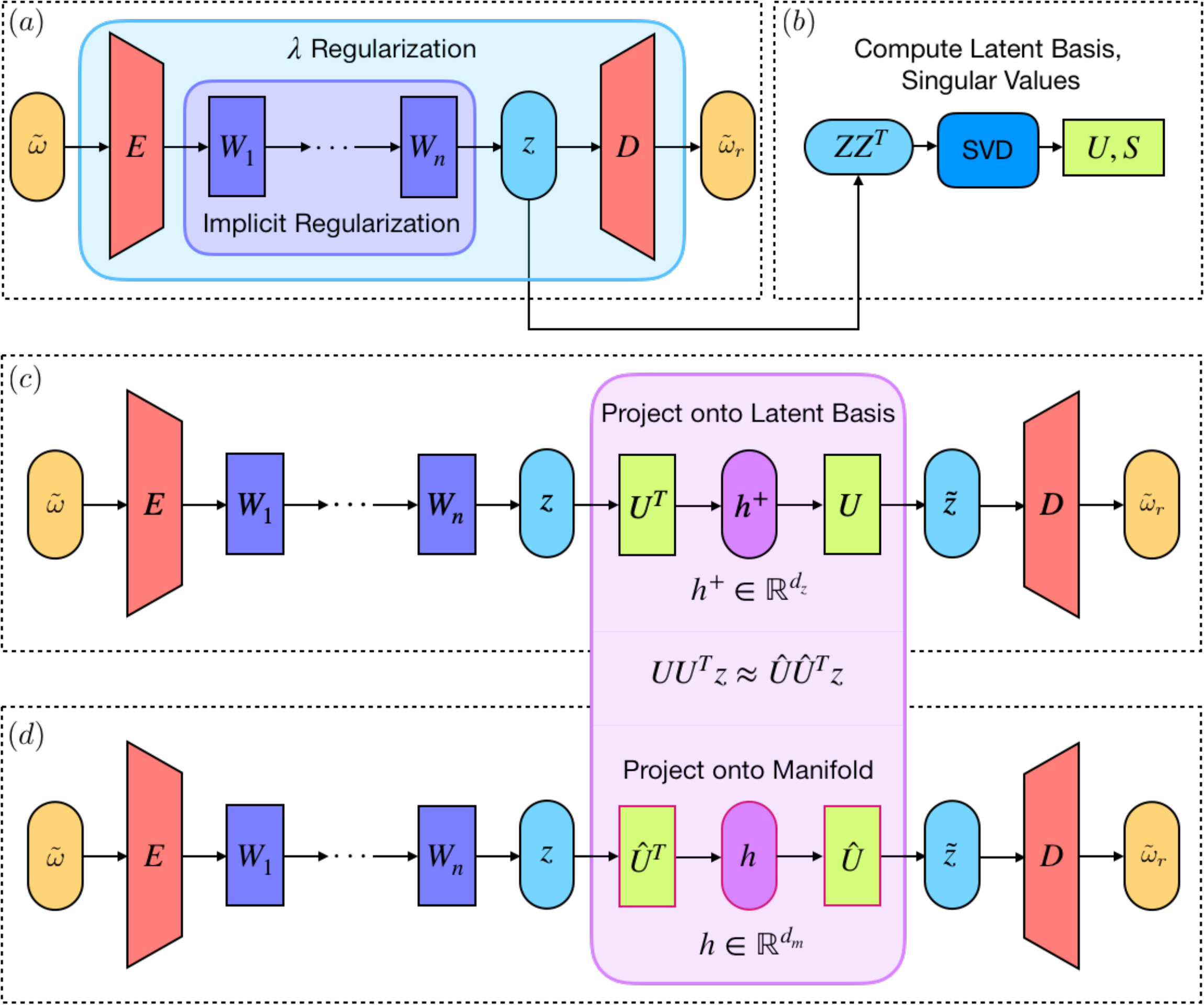}
	\caption{Implicit Rank Minimizing autoencoder with weight decay (IRMAE-WD) framework: a) network architecture with regularization mechanisms, b) singular value decomposition of the covariance of the learned latent data representation $Z$, c) projection of latent variables onto manifold coordinates d) isolated projection of latent variables onto manifold coordinates. Image reproduced with permission from \cite{Zeng.2024.10.1088/2632-2153/ad4ba5}.}
	\label{IRMAE_frame}
\end{figure}

Fig. \ref{IRMAE_frame}\textcolor{blue}{a} shows the IRMAE-WD framework. The encoder, denoted by $z=E\left(\tilde{\omega} ; \theta_{E}\right)$ reduces the dimension from $N$ to the bottlneck dimension $d_z$. We then include a linear network $\mathcal{W}\left(\cdot ; \theta_{W}\right)$ between the encoder and the decoder that consists of several linear layers (matrix multiplications). Finally, the decoder $\tilde{\omega}_r=D\left(z ; \theta_{D}\right)$ maps back to the full space. An $L_2$ (``weight decay") regularization to the weights is also added, with prefactor $\lambda$. The loss function for this architecture is
\begin{equation}
\mathcal{L}\left(\tilde{\omega} ; \theta_{E}, \theta_{W}, \theta_{D}\right)=\left\langle\left\|\tilde{\omega}-D\left(\mathcal{W}\left(E\left(\tilde{\omega} ; \theta_{E}\right) ; \theta_{W}\right) ; \theta_{D}\right)\right\|_{2}^{2}\right\rangle+\frac{\lambda}{2}\|\theta\|_{2}^{2}.\label{eq:IRMAEloss}
\end{equation}
where $\langle\cdot\rangle$ is the average over a training batch, $\theta_E$ the weights of the encoder, $\theta_D$ the weights of the decoder, and $\theta_W$ the weights of the linear network. Zeng et al. \cite{Zeng.2024.10.1088/2632-2153/ad4ba5} found dimension estimates for the KSE to be robust to the number of linear layers, $\lambda$, and $d_z$. However, as we will show, there is more variability when selecting these parameters for Kolmogorov flow.  {Nevertheless, we will see that IRMAE-WD yields a robust estimate of the upper bound of $d_\IM$ that proves very useful.}

After training, we perform SVD on the covariance matrix of the encoded data matrix $Z$ to obtain the singular vectors (columns of $U$) and singular values (diagonal elements $\sigma_i$) of $S$ as shown in Fig. \ref{IRMAE_frame}\textcolor{blue}{b}. Projecting $z$ onto $\hat{U}^T$ gives an orthogonal representation $h^+=U^Tz$ as illustrated in  Fig. \ref{IRMAE_frame}\textcolor{blue}{c}. Then, by choosing only the singular values $\sigma_i$ above some very small threshold (typically $\gtrsim 6$ orders of magnitude smaller than the leading singular values), we may project down to fewer dimensions by projecting onto the corresponding singular vectors, contained in the matrix $\hat{U}$, to yield the low-rank manifold representation $h=\hat{U}z$  (Fig. \ref{IRMAE_frame}\textcolor{blue}{d}). We refer to Table \ref{tablenn} for details on the architecture.

\subsection{Time evolution of pattern with neural ODEs}\label{sec:time_evol}

In the previous two steps we first mapped our data to the fundamental chart, allowing us to represent the state with $\tilde{\omega}$, $\mathcal{I}$, and $\phi_x$. Then we reduced the dimension of $\tilde{\omega}$ by mapping it to the manifold coordinates $h$. Next, we need to find a method to evolve $h$, $\mathcal{I}$, and $\phi$ through time.
To forecast $h(t)$ in time we use the neural ODE (NODE) framework\cite{linot2022data, linot2023stabilized, chen2018neural}, in particular a stabilized version  proposed by Linot et.\ al. \cite{linot2023stabilized} where the dynamics on the manifold in the fundamental chart are described by the equation
\begin{equation}
\frac{d h}{d t}=g_h(h)-Ah.
\end{equation}
Here $A$ is chosen to have a stabilizing effect that keeps solutions from blowing up. We can define this parameter as $A=\text{diag}(\kappa [\text{std}(h)$]) where $\kappa=0.1$ is multiplied by the element-wise standard deviation of $h$. The value of $\kappa$ can be changed, but we see good prediction for the selected value. The equation is integrated to estimate $h_r(t+\tau)$ as 
\begin{equation}
h_r\left(t+\tau\right)=h\left(t\right)+\int_{t}^{t+\tau} (g_h\left(h(t') ; \theta_{t}\right)-Ah) d t'
\end{equation}
where $\theta_t$ are the weights of the NN $g_h$, refer to Table \ref{tablenn}, which are determined by minimizing the loss 
\begin{equation} \label{eq:NODEh}
\mathcal{L}_{\text{NODE}}\left(h ; \theta_{t} \right)=\left\langle\left\|h(t+\tau)-h_r(t+\tau)\right\|_{2}^{2}\right\rangle.
\end{equation}

\MDGrevise{While in the present study we work with clean data from numerical simulations, the methods described here can be used in the presence of noise.   In some cases, the dimension reduction step itself can serve as a noise filter. For example, linear dimension reduction with principal components analysis can be used as a preprocessing step to filter out noise \cite{Young.2023.10.1007/s00397-023-01412-0}. Even in the presence of moderate noise, neural ODEs have been shown to perform well  without changes to the training procedure \cite{chen2018neural,Kidger2021}. If the noise is severe, other steps can be taken, such as smoothing the data with a temporal filter  or adding additional regularization to the neural ODE training   \cite{Kidger2021,Finlay2020,Kelly2020}.}

\MDGrevise{Many studies have used variants of the Koopman operator formalism for data-driven time-evolution modeling \cite{Mezić.2021.10.1090/noti2306,Brunton.2022.10.1137/21m1401243}. A widely used implementation of this approach has recently been shown to be equivalent to a NODE representation \cite{Buzhardt.2024}.}

To train the NODE model we first gather $\omega (t)$ and $\omega (t+\tau)$. We then map to the fundamental chart in pairs. As an example, if a snapshot $\omega (t)$ lies in $\mathcal{I}=5$, we apply the corresponding discrete operation $\mathscr{R}\mathscr{S}$ such that $\tilde{\omega} (t)$ and thus $h(t)$ fall in the interior of the fundamental chart $\mathcal{I}=0$. The same discrete operation (refer to Table \ref{table_I}), is applied to $\omega (t+\tau)$. This means that $h(t+\tau)$ does not need to fall in the interior. We select $\tau=0.5$ which is a small enough time such that the exterior region is covered by the autoencoder. 

\begin{figure}
	\centering
	\begin{subfigure}{.5\textwidth}
		\centering
		\includegraphics[width=1\linewidth]{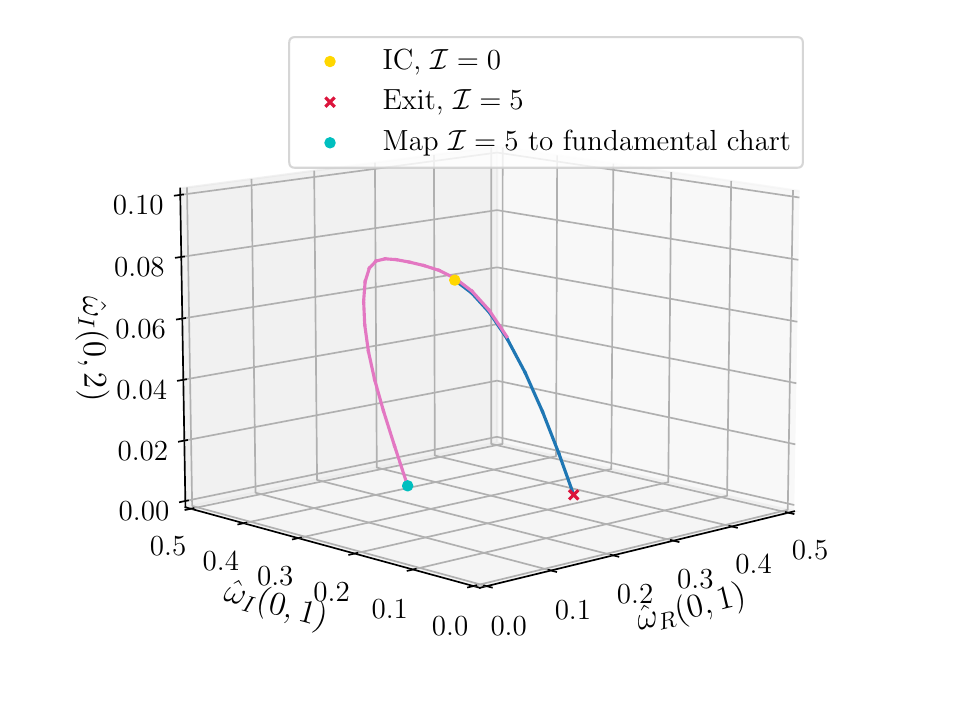}
		\caption{}
		\label{fig:exit_entry_single}
	\end{subfigure}%
        \begin{subfigure}{.5\textwidth}
		\centering
		\includegraphics[width=1\linewidth]{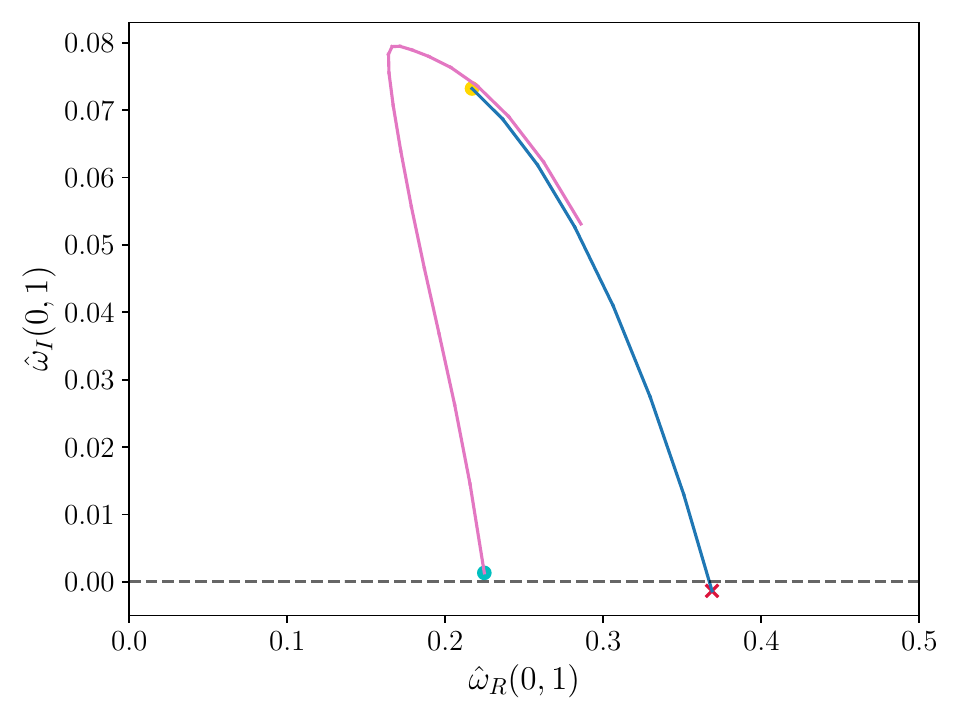}
		\caption{}
		\label{fig:exit_entry}
	\end{subfigure}%
	\caption{ (a) Initial condition (IC) starting at $\mathcal{I}=0$ is evolved and exits into the chart  $\mathcal{I}=5$. This exit is mapped back to $\mathcal{I}=0$ to keep evolving in the fundamental chart. (b) Two-dimensional projection showing the exit from the fundamental chart. Here we selected an IC near the RPO, which is the reason why this trajectory nearly closes on itself.}
	\label{exit_entry_graph}
\end{figure}

\begin{figure}
	\centering
	\includegraphics[width=0.5\linewidth]{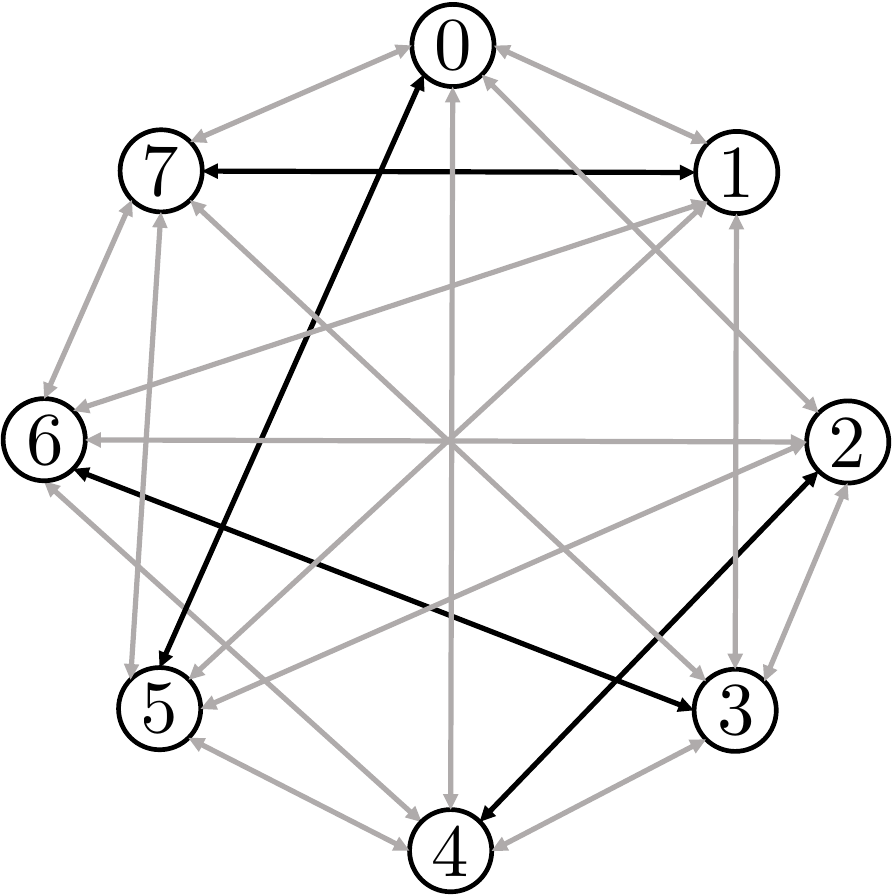}
	\caption{Graph representation of connections between the symmetry subspaces where pairs \{2,4\}, \{0,5\}, \{6,3\}, \{7,1\} correspond to the top and bottom sections of the octants where each unstable RPO lies. The intensity of each connection is related to the probability of the trajectory transitioning to another chart. The probability of staying in the same chart (not included in this depiction) is $\sim 94 \%$, . Darker lines correspond to forward and reverse probabilities of $\sim 5 \%$. Lighter lines correspond to forward and reverse probabilities of $\leq 1 \%$. }
	\label{fig:graph}
\end{figure}

Now that we trained the NODE models we want to evolve trajectories in time. To do this we need to track the pattern $h$ and the indicator $\mathcal{I}$ at every time. We first show an exit and entry in the state space representation in Fig. \ref{exit_entry_graph}. The initial condition (IC) starts in $\mathcal{I}=0$ and exits through the bottom part of the domain. Looking at Fig. \ref{Re13d5_allsym} we see that this corresponds to a transition from $\mathcal{I}=0$ to $\mathcal{I}=5$. Hence, to keep evolving in the fundamental chart we need to apply the corresponding discrete operations to map to $\mathcal{I}=0$ and keep track of the new indicator $\mathcal{I}=5$. Notice that we need to keep track of the new indicator to map back to the full space at the end. This can generalize to a longer trajectory in which case the indicator changes depending on where the trajectory leaves the fundamental chart. The transitions between indicators corresponding to the different charts are shown in Fig. \ref{fig:graph}. Here we see a graph representation of the connections between the 8 symmetry charts for a trajectory of $10^5$ snapshots. The intensity of each connection is related to the probability of the trajectory transitioning to another chart. The four darker lines correspond to the shadowing of the RPOs and the lighter lines are related to the bursting events. Now that we know how the indicators change we can evolve initial conditions with the NODE models to obtain predicted trajectories in $h$.


\begin{figure}
	\centering
	\includegraphics[width=1\linewidth]{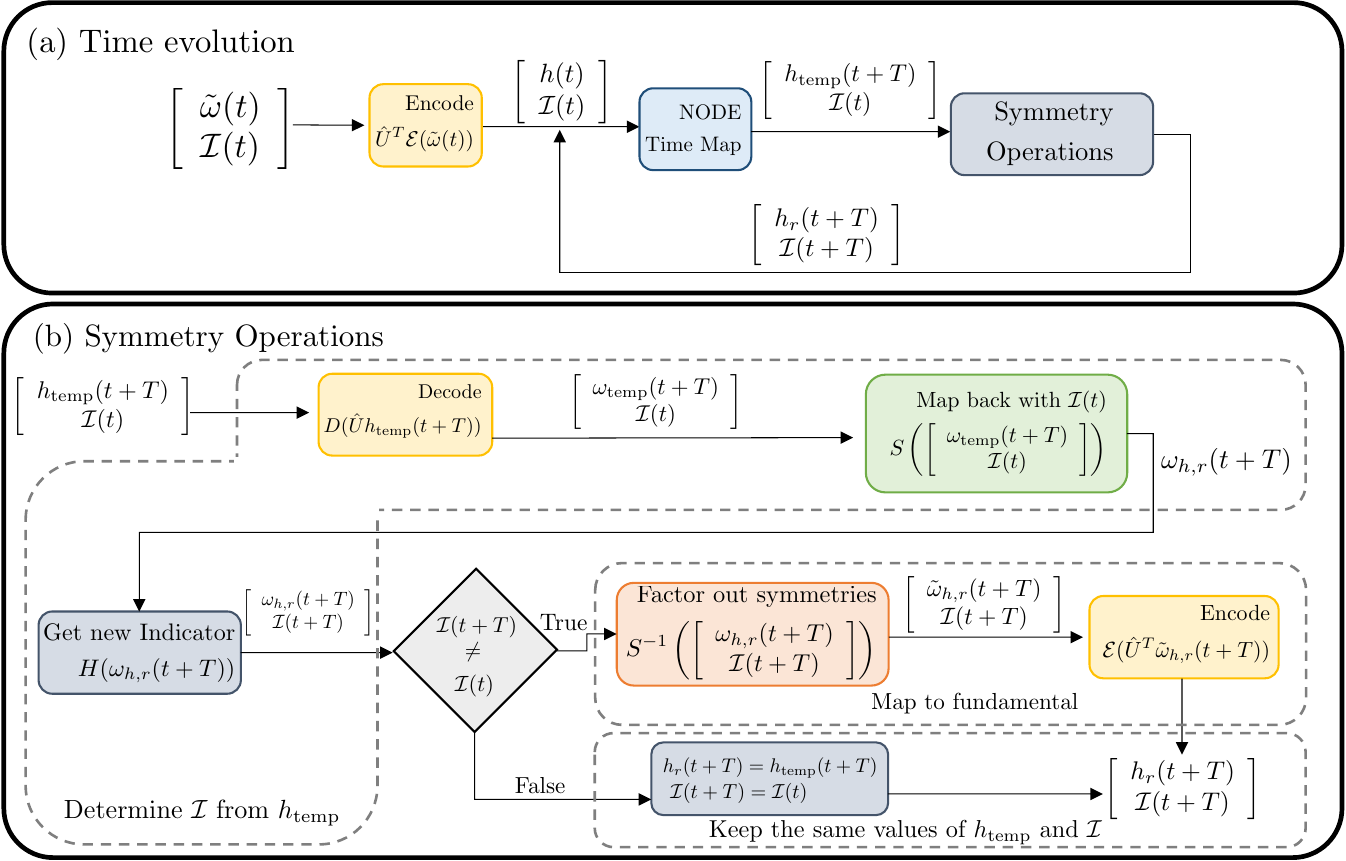}
	\caption{(a) Time evolution of $h(t)$ with NODE and (b) symmetry operation check to map back to the fundamental domain if needed.}
	\label{fig:method}
\end{figure}

Here we discuss the detailed operations for evolving trajectories in the fundamental chart with the NODE models. Fig. \ref{fig:method}\textcolor{blue}{a} summarizes the methodology for evolving in time. An initial condition in the fundamental domain  is first encoded and projected with $\hat{U}$ to obtain $h(t)$. Then the NODE maps a trajectory forward $T$ time units to yield $h_\text{temp}(t+T)$ (note the indicator does not change in this step). We select $T=\tau$. After this, we perform the appropriate symmetry operations, detailed below, to find $h_r(t+T)$ and update the new indicator $\mathcal{I}(t+T)$. This is repeated to continue evolving forward in time. In Fig. \ref{fig:method}\textcolor{blue}{b} we present the method for performing these symmetry operations.


There are two main steps in applying the symmetry operations: 1) determining $\mathcal{I}$ from $h_\text{temp}$, and 2) getting the values of $h_r(t+T)$ and $\mathcal{I}(t+T)$. Step 1 is simply a classification problem where we wish to find a function that classifies new values of $h_\text{temp}$ as either lying within or outside the fundamental domain. In Floryan \& Graham \cite{floryan2021charts} this was performed by identifying the nearest neighbor of the training data in the manifold coordinates and classifying $h_\text{temp}$ with the same label. However, any classification technique can work -- for example, support vector machines (SVM) \cite{boser1992training} also worked well at this task. In the case of our specific problem, we found the fastest method was to map $h_\text{temp}$ to the ambient space $\omega_\text{temp}$ at every step and calculate $\hat{\omega}_R(0, 1)$, $\hat{\omega}_I(0, 1)$, $\hat{\omega}_I(0, 2)$ to verify if $h_\text{temp}$ lies in the fundamental domain. Step 2 involves updating the manifold coordinates to get $h_r(t+T)$. If the indicator changes, symmetries are factored out with $\mathcal{I}(t+T)$ and the snapshot is encoded to get the new $h_r(t+T)$. If the indicator stays the same $h_r(t+T)=h_\text{temp}(t+T)$ and $\mathcal{I}(t+T)=\mathcal{I}(t)$.


\subsection{Time evolution of phase with neural ODEs}\label{sec:time_evol_phase}

We also wish to time-evolve the phase $\phi_x(t)$. The dynamics of the phase only depend upon the fundamental-chart vorticity pattern $h$ and the indicator $\mathcal{I}$, so we can describe them with 
\begin{equation} \label{eq:nodephase}
	\dfrac{d\phi_x}{dt}=g_\mathcal{I}(\mathcal{I})g_\phi(h;\theta_\phi).
\end{equation} 
The term $g_\mathcal{I}(\mathcal{I})$ equals 1 if an even number of discrete operations map $h$ back to the fundamental domain, or -1 if an odd number of discrete operations map $h$ back to the fundamental domain. To understand how the signs of Equation \ref{eq:nodephase} change consider the effect of discrete symmetry operations on the phase calculation $\phi_x=\operatorname{atan} 2\left\{\hat{\omega}_I(1, 0), \hat{\omega}_R(1, 0)\right\}$. Applying a discrete symmetry operation on a snapshot changes the phase variable to either 

\begin{equation}  \label{eq:phaseoper}
	\phi_{x,\mathscr{S}}=\mathscr{S} \phi_x=\operatorname{atan} 2\left\{\hat{\omega}_I(1, 0), -\hat{\omega}_R(1, 0)\right\}=\pi-\phi_x,
\end{equation}
or
\begin{equation}  \label{eq:phaseoper2}
	\phi_{x,\mathscr{R}}=\mathscr{R} \phi_x=\operatorname{atan} 2\left\{-\hat{\omega}_I(1, 0), \hat{\omega}_R(1, 0)\right\}=-\phi_x.
\end{equation}
Then, by simply taking the time derivative, we see that $d\phi_{x,\mathscr{S}}/dt=-d\phi_x/dt$ and $d\phi_{x,\mathscr{R}}/dt=-d\phi_x/dt$. Hence the operation of a discrete symmetry (rotation and shift-reflect) changes the sign. Now, we train $g_\phi$ by fixing the NODE $g_h$ to evolve $h$ forward in time, and use Equation \ref{eq:nodephase} to make predictions $\phi_{x,r}$. We update the parameters of $g_\phi$ to minimize the loss
\begin{equation}
\mathcal{L}_{\phi,\text{NODE}}\left(\phi_{x} ; \theta_{\phi} \right)=\left\langle\left\|\phi_x(t+\tau)-\phi_{x,r}(t+\tau)\right\|_{2}^{2}\right\rangle,
\end{equation}
and refer to Table \ref{tablenn} for details on the NODE architecture.

\begin{table}[h]
\caption{Neural network layer dimensions and activations used in each layer. Sigmoid functions are denoted `S'. Number of linear layers $W_i$ can be varied.}
$$
\begin{array}{lccc}\hline \hline & \text { Function } & \text { Shape } & \text { Activation } \\ \hline \text { Encoder } & E & 1024: 2048:256:40 & \text { ReLU:ReLU:Linear } \\ \text { Decoder } & D & 40: 256: 2048: 1024 & \text { ReLU:ReLU:Linear } \\ \text { Linear Net } & \mathcal{W}=W_1W_2... & 40:40 & \text {Linear } \\ \text { Pattern NODE } & g_h & d_{m}: 500: 500:500: d_{m} & \text { S:S:S:Linear } \\ \text { Phase NODE } & g_\phi & d_{m}: 500:500:500: 1 & \text { S:S:S:Linear } \\ \hline \hline\end{array}
$$
\label{tablenn}
\end{table}

\section{Results} \label{sec:Results}

We present results for the chaotic case $\Rey = 14.4$, whose trajectories sample all eight charts introduced above\MDGrevise{, as well as exhibiting a chaotically drifting phase and thus sampling all values of $\phi_x$}. 
\MDGrevise{We focus on the comparative performance of approaches that do not exploit symmetry and those that do.}
First we illustrate the effect of training data size on the autoencoder performance when considering the fundamental chart data, as opposed to using the original and phase-aligned data. These results are shown in Section \ref{sec:ae_datasize}. {We then  use IRMAE-WD to estimate the minimal dimension needed to represent the data.} We summarize these results in Section \ref{sec:ae_sweep}. The time evolution model performance results are shown in Section \ref{sec:timeresults}. Here we confirm the equivariance of the model,  (Section \ref{sec:Equivariance}) then show the performance for short-time tracking, long-time statistics, and phase evolution (Sections \ref{sec:short_time},  \ref{sec:long_time}, and \ref{sec:phasepred} respectively).




\subsection{Dimension reduction with IRMAE-WD}\label{sec:ae_MSE_results}

To train our IRMAE-WD models, we minimize the loss $\mathcal{L}$ in Eq.~\ref{eq:IRMAEloss} via
stochastic gradient descent. Following Zeng et al.~\cite{Zeng.2024.10.1088/2632-2153/ad4ba5},  we use the AdamW optimizer, which decouples weight decay from the adaptive gradient update and helps avoid the issue of weights with larger gradient amplitudes being regularized disproportionately, as observed in Adam \cite{loshchilov2017decoupled}.  All models were trained for a total of 1000 epochs and with a learning rate of $10^{-3}$. \MDGrevise{The entire data set consists of a time series of $10^5$ snapshots at time intervals of $\tau= 0.5$ time units.  The first $80,000$ snapshots, or downsampled versions thereof as described below, are the training data, and the last $20,000$ are the test data.  For cases where phase alignment is applied, the original training data snapshots $\omega$ are transformed as described above to $\omega_l$, and where the discrete symmetries are also factored out, they are mapped into the fundamental domain. This is then extended to form the fundamental chart; the data points in this chart are denoted $\tilde{\omega}.$ The operations for performing these transformations are described in Sec.\ \ref{sec:prepros}.}

Before training the AEs, the mean is subtracted and the data is divided by its standard deviation. \MDGrevise{For each data set ($\omega$, $\omega_l$, and $\tilde{\omega}$ cases, with varying numbers of samples as described below),}  Three models are trained for each case of varying number of linear layers $L=0,4,6,8,10$ and weight decay values  $\lambda=10^{-4},10^{-6},10^{-8}$.  \MDGrevise{That is, for each data set we train 45 separate models.}
For all autoencoders, we select a bottleneck dimension $d_z=40$, which is significantly higher than the manifold dimension expected based on our previous work \cite{perezde2023data}.

\MDGrevise{In Sec.~\ref{sec:ae_datasize}, we examine the effect of symmetry reduction on the autoencoder error as we change the size of the test data set. In Sec.~\ref{sec:ae_sweep}, we keep the data size constant and examine the effect of symmetry reduction on the estimated manifold dimension. In both cases, we will see that incorporating symmetry has a substantial benefit. }


\subsubsection{Effect of training data size on performance}\label{sec:ae_datasize}


A major benefit of mapping data to the fundamental chart is that it results in eightfold denser sampling within that chart, as shown \MDGrevise{by comparison of Figs.~\ref{chart_blowupa} and }Fig.~\ref{fig:3dproj_14d4_colored_ppp}. We also see that all the data is in a much smaller part of state space, and only that part of state space needs to be represented. In this section, we test to what extent this increased density \MDGrevise{improves the autoencoder performance for a given set of training data. Using the full training data (80,000 snapshots at time intervals $\tau=0.5$), as well as reduced sets obtained by downsampling the training data set (increasing the time interval between snapshots),} we train IRMAE-WD models with $L=4$ and $\lambda=10^{-4}$.  This is done for the original data ($\omega$), the phase-aligned data ($\omega_l$), and the data mapped to the fundamental domain ($\tilde{\omega}$). 

\begin{figure}
	\centering
	\includegraphics[width=0.6\linewidth]{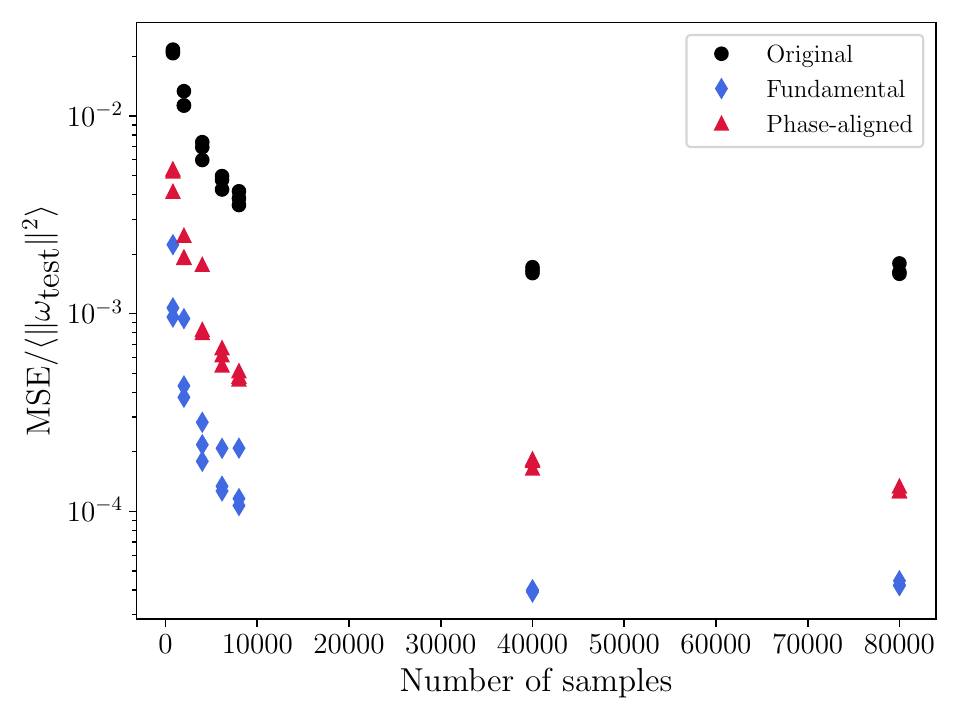}
	\caption{Normalized autoencoder test MSE  vs.~training data size, $\text{Re}=14.4$. The parameters used are $L=4$ and $\lambda=10^{-4}$, and three trials are considered for each case. }
	\label{IRMAE_result}
\end{figure}

In Fig.\ \ref{IRMAE_result} we show the test MSE of the IRMAE-WD models, normalized based on the mean squared vorticity for the test data ($\omega_\mathrm{test}$), as we vary the amount of training data. 
As can be seen in Fig.\ \ref{IRMAE_result}, removing phase results in an order of magnitude improvement in the normalized MSE relative to the original data, and mapping to the fundamental results in nearly another order of magnitude improvement. This this drastic improvement in performance allows us to use far less data. For example, \MDGrevise{by comparing the blue diamonds to the black circles, we see that a model trained with only} 800 snapshots in the fundamental domain performs nearly as well as one trained with $80,000$ snapshots in the original state space. Similarly, $8,000$ snapshots in the fundamental case outperforms $80,000$ snapshots in the phase-aligned case. \ALrevise{We only show the performance for the IRMAE-WD architecture, but our approach only modifies the input data. This means it can be applied to any autoencoder architecture.}


\subsubsection{Effect of symmetry reduction and autoencoder parameters on manifold dimension estimate}\label{sec:ae_sweep}

\begin{figure}
	\centering
	\includegraphics[width=0.6\linewidth]{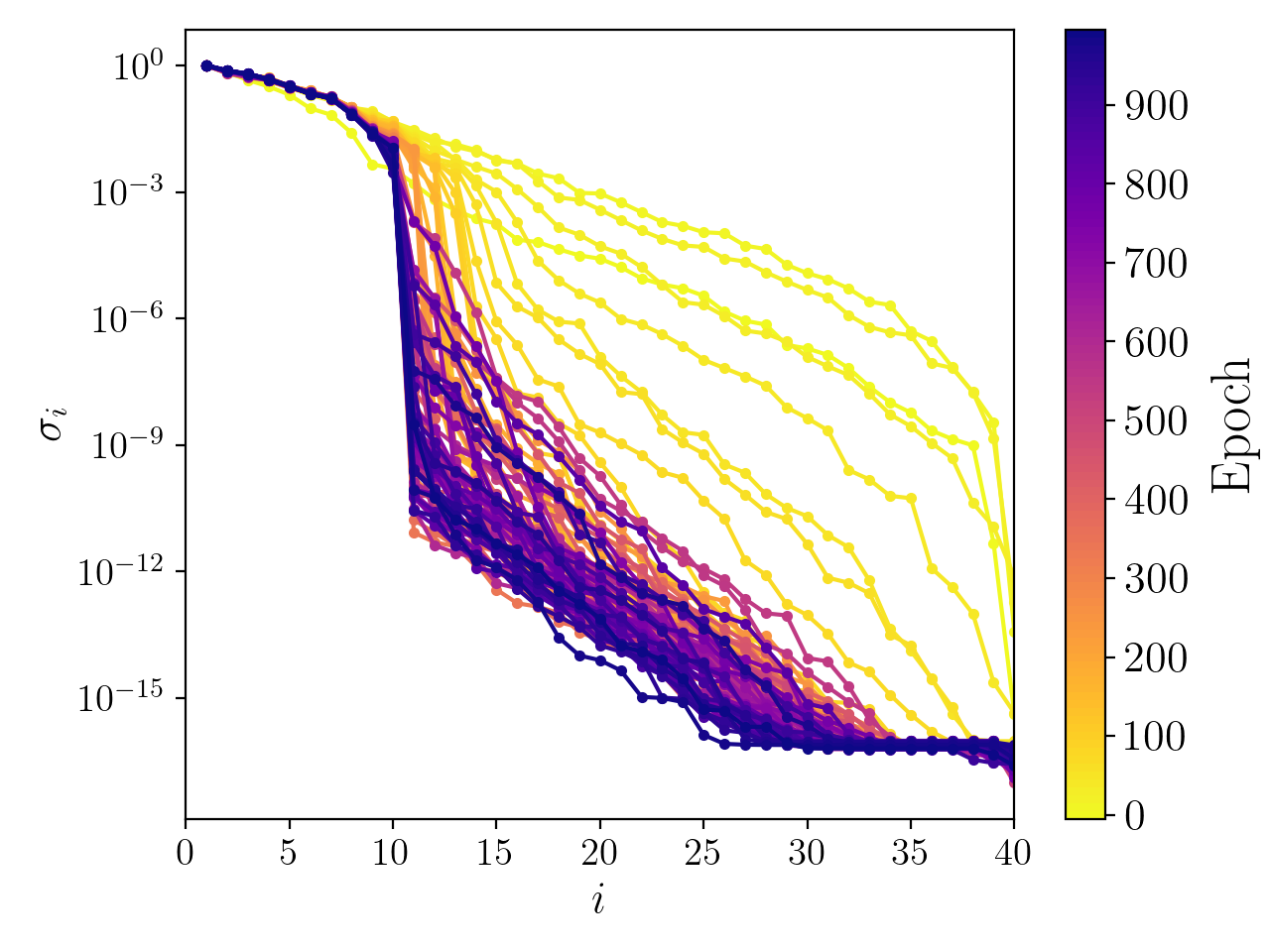}
	\caption{Evolution of the spectrum of singular values \MDGrevise{$\sigma_i$ (ordered by magnitude)} of the covariance matrix of the encoded test data $ZZ^T$ during training of an IRMAE-WD model on the symmetry-reduced data, with $L=4$ and $\lambda=10^{-4}$. \MDGrevise{Color indicates training epoch. } Here the drop happens at $d_h=10$.} 
	\label{svd_dyn}
\end{figure}

Here we study the effects on the dimension estimates  when varying $L$ and $\lambda$ for the original, phase-aligned, and fundamental cases. The full training data set of 80,000 snapshots is used.  As discussed in Section \ref{sec:AE} we perform SVD on the covariance matrix of the encoded data $ZZ^T$ to obtain $U$ and truncate to obtain $h=\hat{U}^Tz$. Zeng et al. \cite{Zeng.2024.10.1088/2632-2153/ad4ba5} {showed that structuring an autoencoder with linear layers and using weight decay causes the latent space to become low-dimensional through training.} In Fig.\ \ref{svd_dyn}, we show an example of the evolution of these singular values through training, using data in the fundamental chart.  As the model trains for longer times, the trailing singular values tend towards zero. These can be truncated without any loss of information. For most cases, this drop is drastic ($\sim 10$ orders of magnitude) and a threshold can be defined to select how many singular values to keep (i.e.~to select the number of dimensions).

 \begin{figure}
	\centering
	\includegraphics[width=0.6\linewidth]{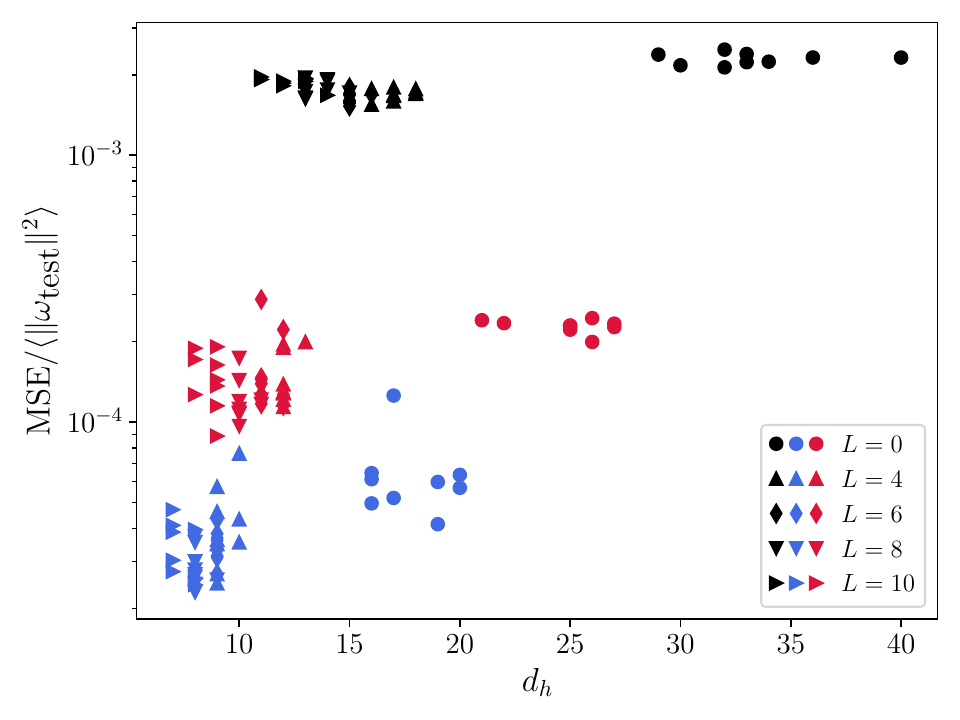}
	\caption{Normalized MSE vs.~dimension $d_h$ given by the spectral gap of the singular values for the original (black), phase-aligned (red), and fundamental chart (blue) cases. Each case contains three trials of combinations of parameters $L=0,4,6,8,10$ and $\lambda=10^{-4},10^{-6},10^{-8}$.}
	\label{fig:MSE_sweep_all}
\end{figure}

In Fig.\ \ref{fig:MSE_sweep_all} we plot normalized MSE vs.~estimate of dimension for the original (black), phase-aligned (red), and fundamental chart (blue) data as we vary the number of linear layers $L$ and the weight decay parameter $\lambda$. For each case, there are two clear clusters -- the cluster with higher dimension corresponds to $L=0$ and the other cluster has linear layers, $L>0$. {This happens because in the absence of linear layers there is no mechanism to drive the rank to a minimal value.} When we factor out symmetries, the MSE decreases (as shown above), and the dimension estimates become less spread, with ranges of $d_h=7-10$ for fundamental, $d_h=8-13$ for phase-aligned, and $d_h=11-18$ for original. This narrowing of the distribution likely happens due to the dense coverage of the state space in the fundamental chart, enabling the autoencoder to better capture the shape of the manifold. The dimension estimate range from the fundamental chart is in good agreement with our previous observations of $d_h=9$ in \citep{perezde2023data}. In the following analysis, we select a conservative estimate of the dimension, which appears to be $d_h=10$ from the fundamental chart data.



\subsection{Time evolution}\label{sec:timeresults}

To learn our NODE models we first train $g_h$ by minimizing Equation \ref{eq:NODEh} for $\mathcal{L}_{\text{NODE}}$, and then we fix $g_h$ and train $g_\phi$ by minimizing Equation \ref{eq:nodephase} ($\mathcal{L}_{\phi,\text{NODE}}$) via the
Adam optimizer. We train for a total of 40000 epochs and a learning rate scheduler that drops from $10^{-3}$ to $10^{-4}$ at epoch number 13334 (1/3 into training) and from $10^{-4}$ to $10^{-5}$ at epoch number 26667 (2/3 into training). As previously discussed we use $\tau=0.5$ which ensures the trajectory spends some time steps in an overlap region as it moves from chart to chart, so we can learn the dynamics there.


\subsubsection{Equivariance}\label{sec:Equivariance}

\begin{figure}
	\centering
	\includegraphics[width=0.6\linewidth]{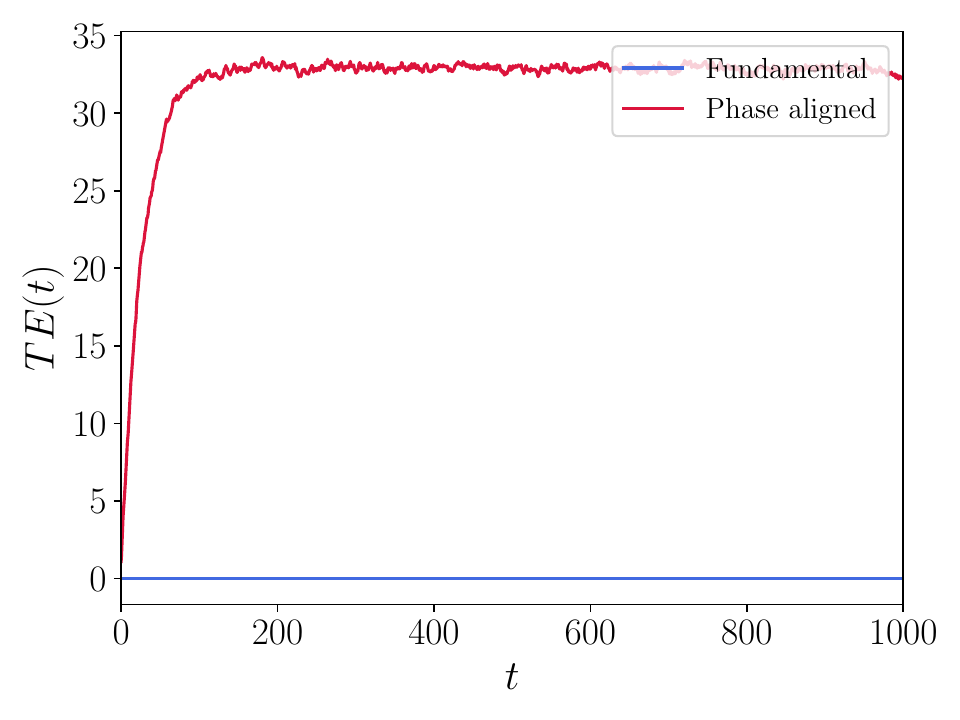}
	\caption{$TE(t)$ vs.~$t$ for Fundamental and Phase-aligned case where $TE(t)$ corresponds to the error calculation between predicted NODE trajectories, from a model with $d_h=10$, from initial conditions that are related by symmetry transformations.}
	\label{equiv_error}
\end{figure}

The results we obtain with this framework should be equivariant with respect to initial conditions. This means that after we apply any of the symmetry operations described above to an initial condition, the resulting trajectory from the original initial condition and the new initial condition must be equivalent up to this symmetry operation.
Here we show that our methodology retains equivariance. We select the IRMAE-WD models with the lowest MSEs at a dimension of $d_h=10$, which is 
a conservative dimension estimates for the fundamental case, for both the phase-aligned and fundamental model. We then sample 1000 initial conditions separated by 10 time units such that we cover different regions of state space. For every initial condition, we apply all the discrete symmetry operations in the original Fourier space, mapping the data into every octant. Then we evolve these initial conditions forward 1000 time units with the DManD model, with and without symmetry charting. To test for equivariance, we compute the trajectory error $TE$ between predicted trajectories as follows
\begin{equation}
TE(t)=\frac{1}{100 \times 7}\sum_{i=1}^{100} \sum_{j=1}^{7}  \| \tilde{\omega}_{i,j}(t)-\tilde{\omega}_{i,j=0}(t) \|,
\end{equation}
where $i$ corresponds to the trajectory number, and $j$ to the  initial chart (i.e.~$j=0$ corresponds to the original initial condition of the $i$th trajectory). Fig. \ref{equiv_error} presents this error through time for our symmetry charting method (blue) and for models with only phase-alignment (red). 
As expected, the time integration from the model trained in the fundamental domain satisfies equivariance perfectly with a $TE=0$. The phase-aligned curve does not satisfy equivariance, and we see that trajectories diverge fairly quickly, leading to substantial error. 

\begin{figure}
	\centering
	\begin{subfigure}{.5\textwidth}
		\centering
		\includegraphics[width=1\linewidth]{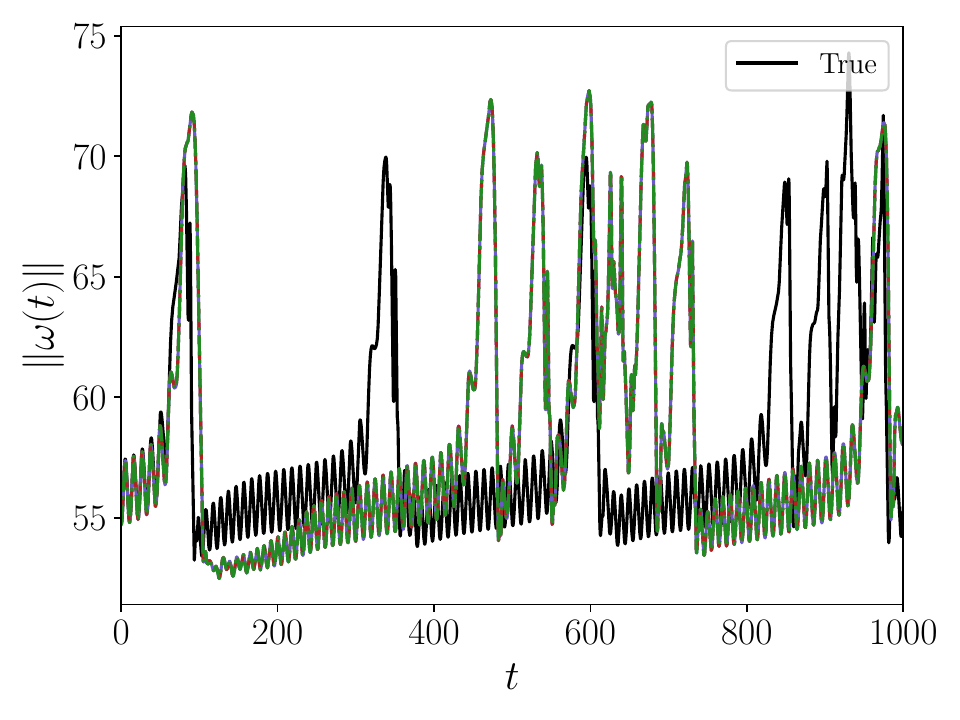}
		\caption{}
		\label{fig:enstr_Fund}
	\end{subfigure}%
	\begin{subfigure}{.5\textwidth}
		\centering
		\includegraphics[width=1\linewidth]{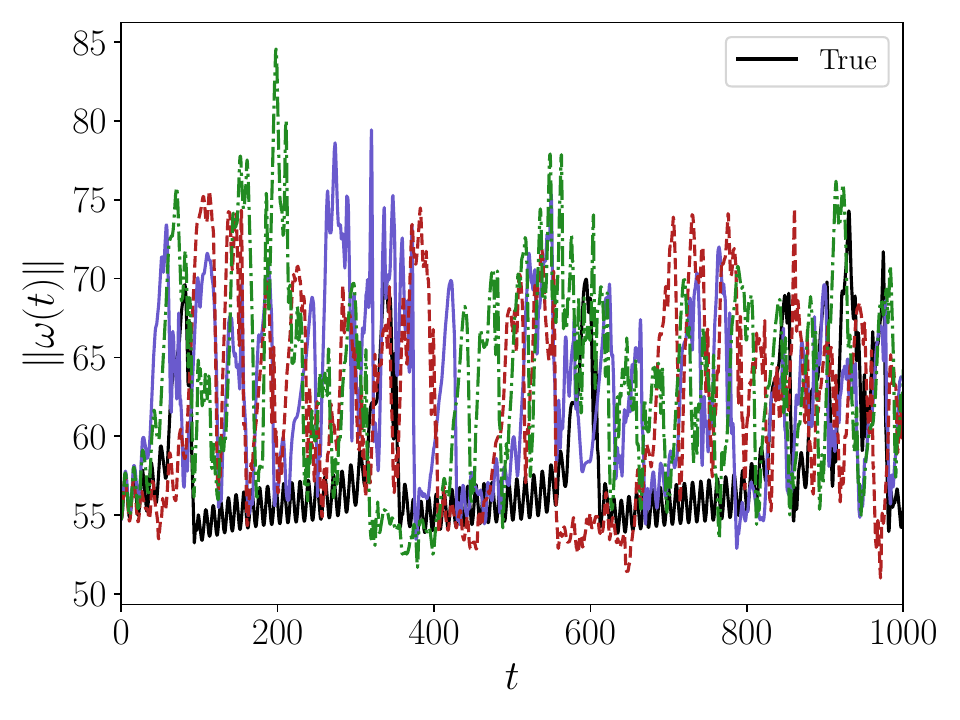}
		\caption{}
		\label{fig:enstr_Phase}
	\end{subfigure}
	\caption{ $\| \omega (t) \| $ vs.~time for initial conditions integrated with the (a) Fundamental and (b) Phase aligned NODE model for $d_h=10$. Colors correspond to initial conditions starting in different fundamental domains (red -- $\ind=0$, blue -- $\ind=1$, and green -- $\ind=2$). In (a), all the different-colored trajectories coincide.}
	\label{fig:enstrophy_all}
\end{figure}

A more severe consequence of not enforcing the discrete symmetries can be seen in trajectory predictions. Fig. \ref{fig:enstrophy_all} depicts  $\| \omega (t) \| $ vs.~$t$ for the true data (shown in black) and the time integration of initial conditions starting in different charts (red -- $\ind=0$, blue -- $\ind=1$, and green -- $\ind=2$) for both fundamental (a) and phase-aligned cases (b). In all cases, the predicted trajectories diverge from the true trajectory after some time -- this is expected for a chaotic system, as explore further in Section \ref{sec:short_time}. However, the symmetry-charted predictions in Fig. \ref{fig:enstr_Fund} exhibit excellent short-time tracking and captures the bursting event that happens around $t=100$. At longer times, the prediction is not quantitatively accurate but still captures the alternation between quiescent and bursting intervals observed in the true data. By contrast, 
the predictions for the phase-aligned model, Fig. \ref{fig:enstr_Phase}, deviate quickly from the true data, and furthermore do not even exhibit intermittency between quiescent and bursting dynamics -- they stay in a bursting regime. Thus the models that do not account for the discrete symmetries do not capture the dynamics correctly, even at a qualitative level. These results reinforce the major advantage of properly accounting for symmetries, as the symmetry charting method does.

\subsubsection{Short-time predictions}\label{sec:short_time}

In this section, we focus on short-time trajectory predictions. The Lyapunov time $t_L$ for this system is approximately $t_L \approx 20$ \cite{inubushi2012covariant}. We take initial conditions of $h(t)$ and evolve for $100$ time units. These are then decoded and compared with the true vorticity snapshots. We consider trajectories with initial conditions starting in the quiescent as well as in the bursting regions. The dynamics at $\text{Re}=14.4$ are characterized
by quiescent intervals where the trajectories are close to RPOs (which are now unstable), punctuated
by heteroclinic-like excursions (bursting) between the RPOs, which are indicated by the intermittent increases of $\| \omega (t) \|$ as observed in Fig. \ref{fig:ens_true_14d4}. The nature of the intermittency of the data makes it challenging to assign either bursting or quiescent labels. To split the initial conditions as quiescent or bursting we use the algorithm discussed in \cite{perezde2023data}.

\begin{figure}
	\centering
	\begin{subfigure}{.5\textwidth}
		\centering
		\includegraphics[width=1\linewidth]{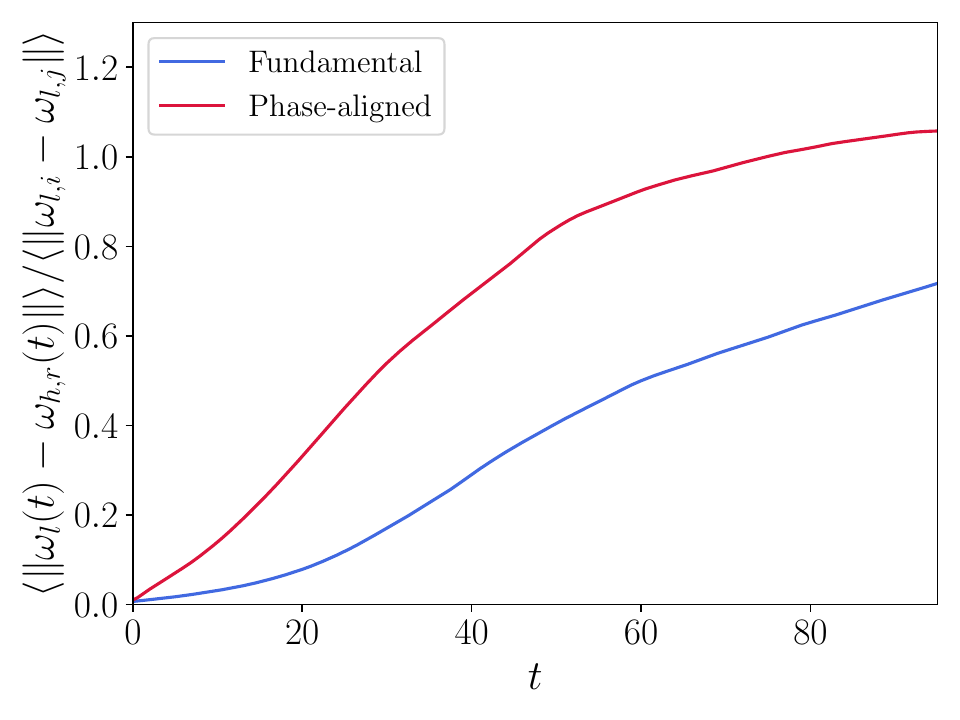}
		\caption{}
		\label{fig:short_track_tot}
	\end{subfigure}%
	\begin{subfigure}{.5\textwidth}
		\centering
		\includegraphics[width=1\linewidth]{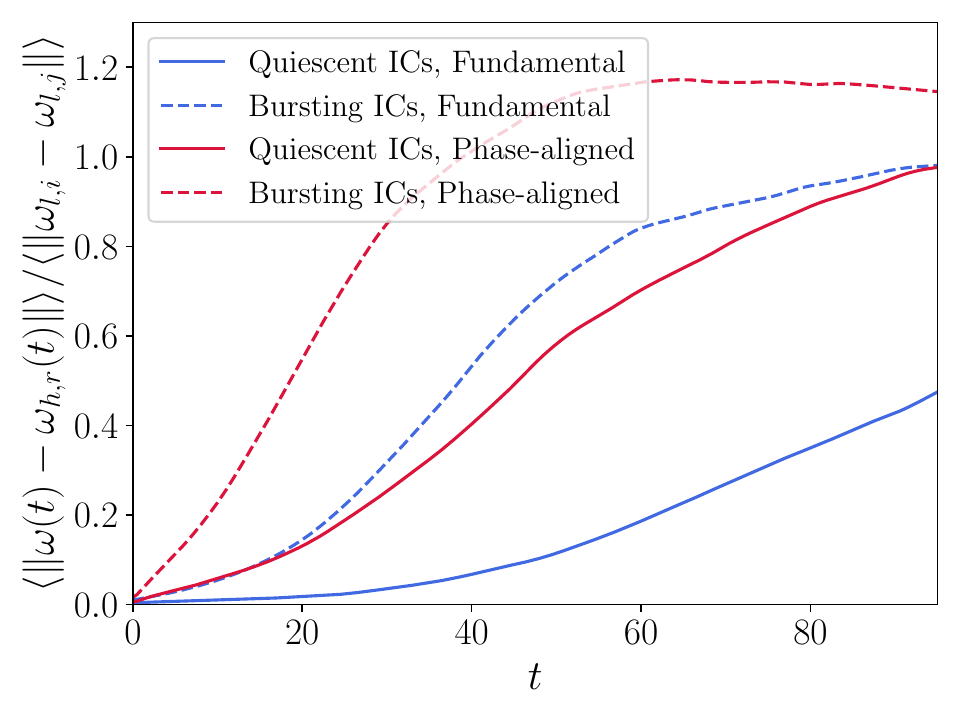}
		\caption{}
		\label{fig:short_track_hibbur}
	\end{subfigure}
	\caption{ Difference between true vorticity evolution and vorticity evolution obtained from the Fundamental and Phase aligned NODEs from $h(t)$ of $d_h=10$, where (a) corresponds to averages over all initial conditions, and (b) corresponds to averages taken over bursting and quiescent initial conditions. }
	\label{fig:short_track}
\end{figure}

We first show in Fig. \ref{fig:short_track_tot} the ensemble-averaged prediction error as a function of time for $10^4$ initial conditions. We use the same models as in the previous section which corresponds to $d_h=10$ for both fundamental and phase-aligned case. The comparison is done with true phase-aligned data, so after obtaining the prediction from the fundamental case we use the indicators to include the symmetries. The error is normalized with random differences of the true data, where $i$ and $j$ correspond to different snapshots. With this normalization, when the curves approach 1 this means that on average the distance between the model and the true data is the same as if we selected random points from the true data. The DManD models using symmetry charting significantly outperform the phase-aligned DManD models. This agrees with Fig. \ref{fig:enstrophy_all} as discussed above. Similar improvement is observed in Fig. \ref{fig:short_track_hibbur}, and can be attributed to the organized (near-RPO) nature of the dynamics in the quiescent region. Also, the dynamics spend more time in this area, so there is more data for the autoencoder to train on. 


\subsubsection{Long-time predictions}\label{sec:long_time}

\begin{figure}
	\centering
	\begin{subfigure}{.5\textwidth}
		\centering
		\includegraphics[width=1\linewidth]{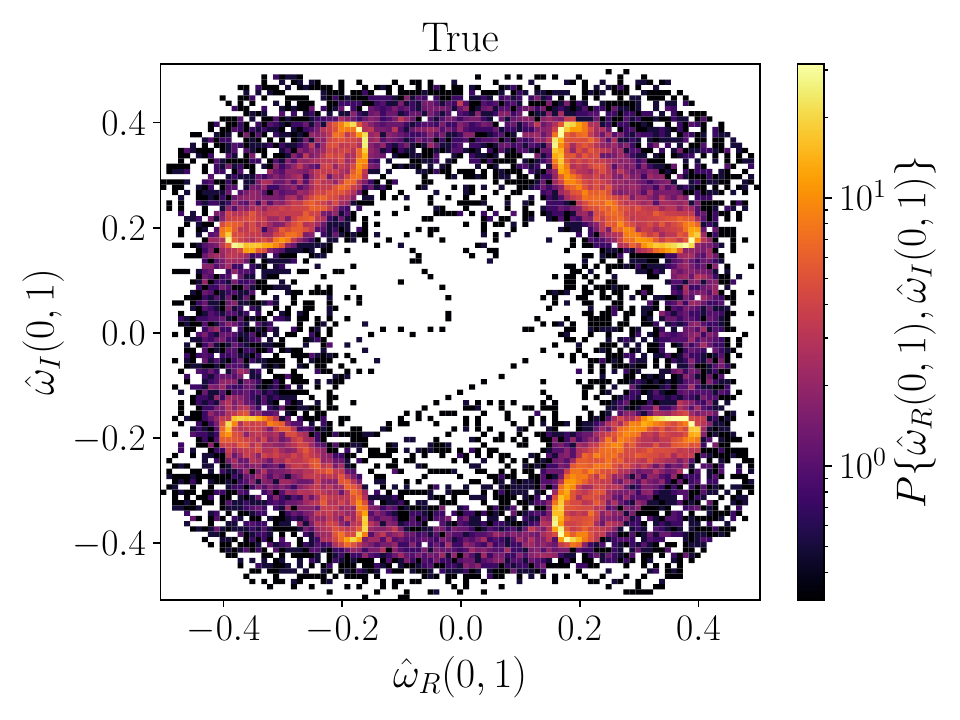}
		\caption{}
		\label{fig:Re_Im_pdf_True}
	\end{subfigure}%
	\begin{subfigure}{.5\textwidth}
		\centering
		\includegraphics[width=1\linewidth]{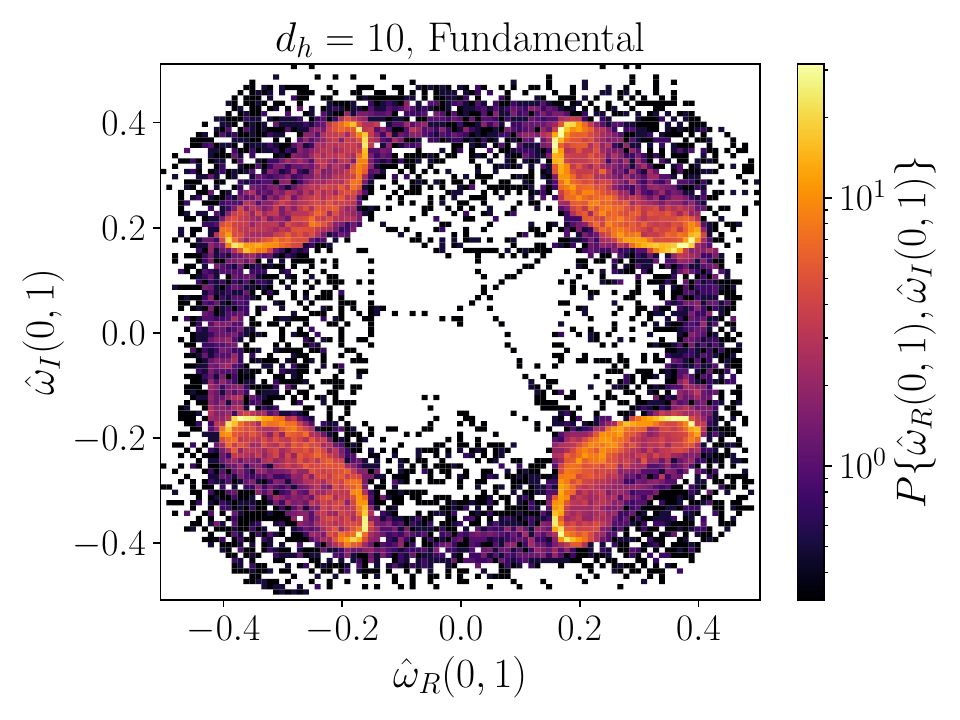}
		\caption{}
		\label{fig:Re_Im_pdf_dh10}
	\end{subfigure}
        \begin{subfigure}{.5\textwidth}
		\centering
		\includegraphics[width=1\linewidth]{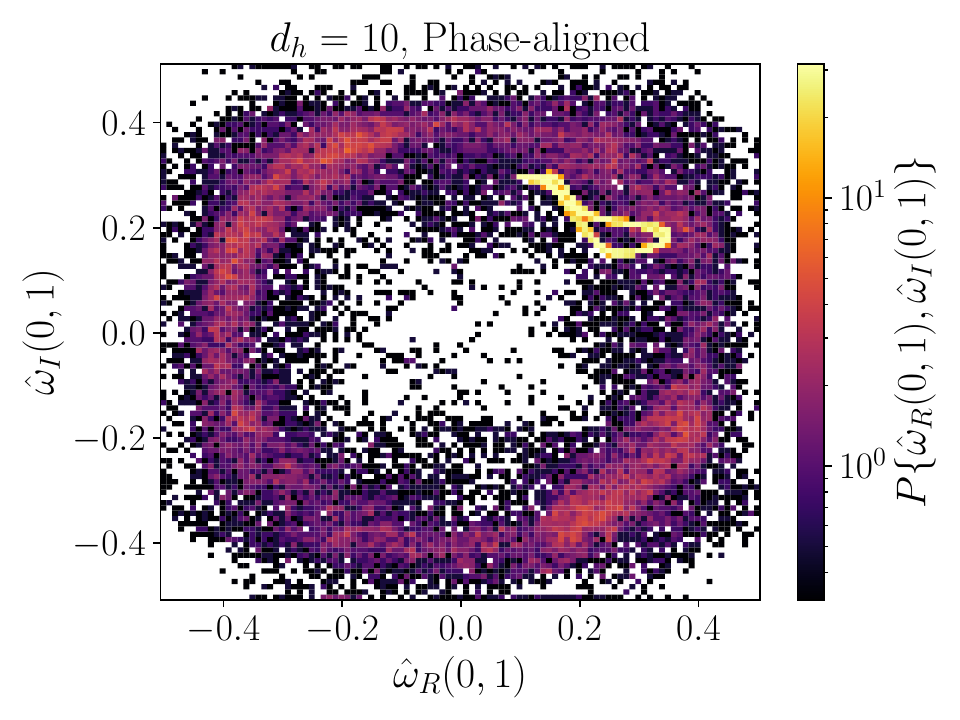}
		\caption{}
		\label{fig:Re_Im_pdf_dh10_Phase}
	\end{subfigure}
	\caption{ Joint PDFs of $\hat{\omega}_{R}(0,1)$-$\hat{\omega}_{I}(0,1)$ of (a) true, and predicted data corresponding to dimension $d_h=10$ from the (b) Fundamental and (c) Phase-aligned models.}
	\label{fig:Re_Im_pdf}
\end{figure}

Now that we demonstrated the short-time predictive capabilities of the model, we next turn to the ability of the model to reconstruct the long-time statistics of the attractor.
For this comparison, we sample $2 \times 10^4$ time units of data every $0.5$ time units for the DNS and the DManD models.
Fig. \ref{fig:Re_Im_pdf} shows the joint probability density function (PDF) of $\text{Re}[a_{0,1}(t)]$ and $\text{Im}[a_{0,1}(t)]$ for true and predicted data from the models with $d_h=10$ for the fundamental and phase-aligned case. 
The true joint PDF (Fig.\ \ref{fig:Re_Im_pdf_True}) and the symmetry charting fundamental joint PDF (Fig.\ \ref{fig:Re_Im_pdf_dh10}) are in excellent agreement. These PDFs both show a strong, equal preference for trajectories to shadow the four unstable RPOs, and lower probabilities between them where the bursting transitions of the RPO regions occur. In contrast, the joint PDF for the phase-aligned model (Fig.\ \ref{fig:Re_Im_pdf_dh10_Phase}) shows that the model samples the space before falling onto an unphysical stable RPO (bright closed curve) near one of the true system's unstable RPOs. Clearly, in this case, the phase-aligned model fails to capture the system's true dynamics. 

\begin{figure}
	\centering
	\begin{subfigure}{.5\textwidth}
		\centering
		\includegraphics[width=1\linewidth]{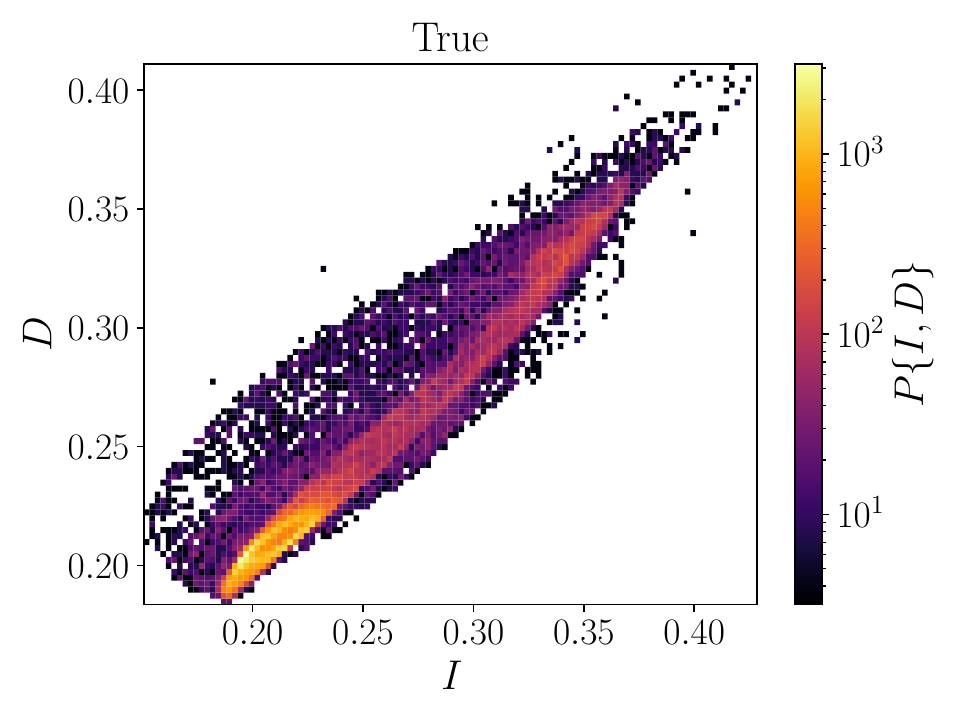}
		\caption{}
		\label{fig:IP_D_pdf_True}
	\end{subfigure}%
	\begin{subfigure}{.5\textwidth}
		\centering
		\includegraphics[width=1\linewidth]{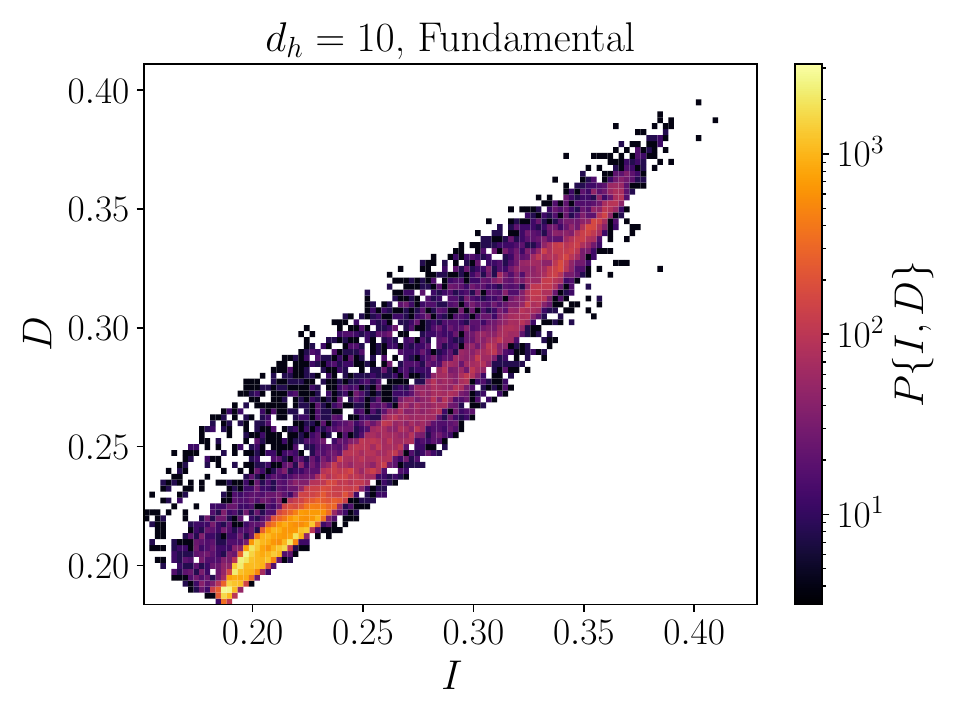}
		\caption{}
		\label{fig:IP_D_pdf_dh10}
	\end{subfigure}
        \begin{subfigure}{.5\textwidth}
		\centering
		\includegraphics[width=1\linewidth]{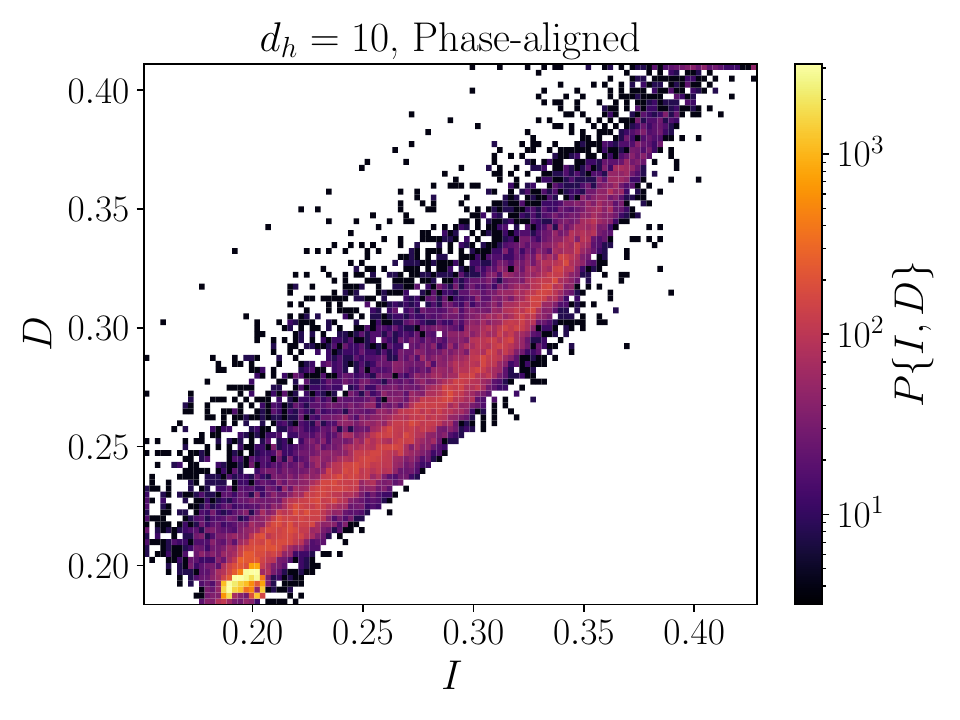}
		\caption{}
		\label{fig:IP_D_pdf_dh10_Phase}
	\end{subfigure}
	\caption{ Joint PDFs of $I$-$D$ of (a) true, and predicted data corresponding to dimension $d_h=10$ from the (b) Fundamental and (c) Phase-aligned models. }
	\label{fig:IP_D_pdf}
\end{figure}

Another important quantity to consider is the ability of the models to capture the energy balance of the system. In Fig. \ref{fig:IP_D_pdf} we show the joint PDF of $I$
and $D$ for the DNS and the same models. Again, the model trained in the fundamental domain closely matches the joint PDF of the true data. The phase-aligned model both underestimates the energy associated with the high probability RPOs, and overestimates the energy associated with the low probability high power input and dissipation events.


\subsubsection{Phase variable prediction}\label{sec:phasepred}

To complete the dynamical picture we predict the phase evolution as given by Equation \ref{eq:nodephase}. We compare the models to the true data by calculating the mean squared displacement (MSD) of the phase,
\begin{equation}
\mbox{MSD} (t)= \langle (\phi_x (t) - \phi_x (0))^2 \rangle
\end{equation}
as was done in our previous work \cite{perezde2023data}. Due to the bursting region, where the direction of phase evolution is essentially randomly reset, at long times the phase $\phi_x$ exhibits random-walk behavior, which the MSD reflects.
We take $420$ initial conditions separated by $15$ time units and use the models of $d_h=10$ to predict $\phi_x(t)$ and calculate the MSD. This is done for the fundamental and phase-aligned models.

 \begin{figure}
	\centering
	\includegraphics[width=0.6\linewidth]{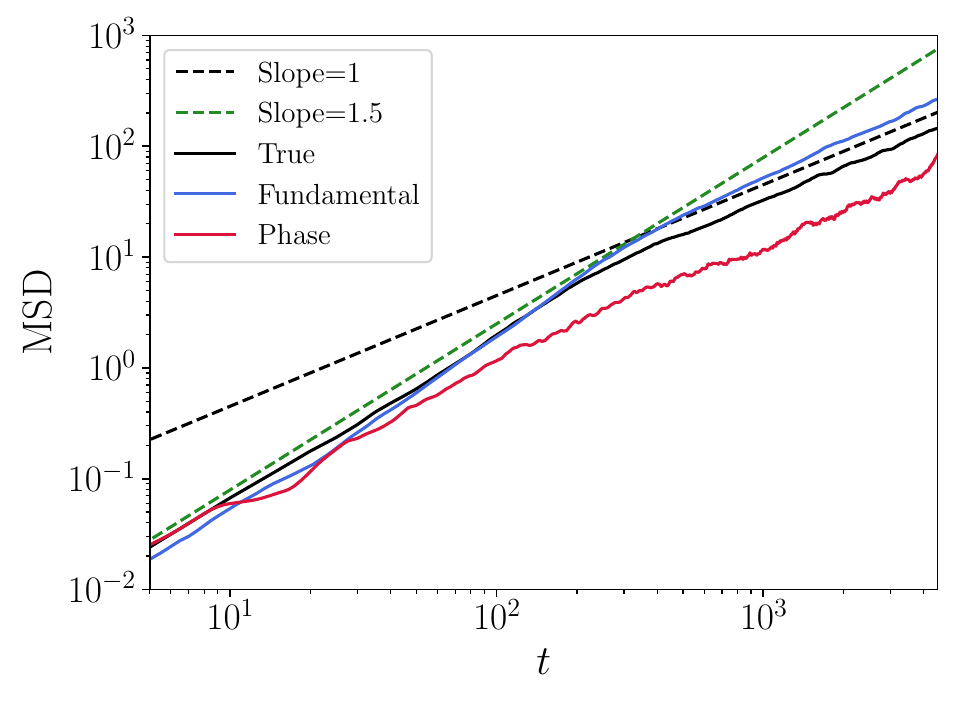}
	\caption{MSD of $\phi_x(t)$ corresponding to models with dimension $d_h=10$.}
	\label{MSD}
\end{figure}

Fig. \ref{MSD} shows the evolution of MSD of true and predicted data. Here the black solid line corresponds to the true data, and the black and green dashed lines serve as references with slopes of $1$ and $1.5$ respectively. There is a change from superdiffusive ($MSD\sim t^{1.5}$) to diffusive ($MSD\sim t$) scaling around $t \approx 200$, which corresponds to the mean duration of the quiescent intervals: i.e., to the average time the trajectories travel along the RPOs before bursting. The fundamental chart model accurately captures both the change in slope and the timing of this change in slope. However, the curve of the phase-aligned model fails to match the true MSD at long times. This happens because the phase-aligned model is unable to predict $h(t)$ correctly at long times as observed in Fig. \ref{fig:enstr_Phase}, hence resulting in poor predictions for $\phi_x(t)$.


\section{Conclusion} \label{sec:Conclusion}

Symmetries appear naturally in dynamical systems and we show here that correctly accounting for them dramatically improves the performance of data-driven models for time evolution on an invariant manifold. 
In this work, we introduce a method that we call symmetry charting and apply it to Kolmogorov flow in a chaotic regime of $\text{Re}=14.4$. This symmetry charting method factors out symmetries so we can train ROMs in a fundamental chart and ensure equivariant trajectories -- we are essentially learning coordinates and dynamics on one region (chart) of the invariant manifold for the long-time dynamics rather than having to learn these for the entire manifold. To do so, we first factor out the symmetries by identifying a set of indicators that differentiate the set of discrete symmetries for the system. Here, Fourier coefficients serve this purpose; we found that the signs of specific Fourier modes could uniquely identify all discrete symmetry operations. {We then use IRMAE-WD, an autoencoder architecture that tends to drive the rank of the latent space covariance of the data to a minimum, to find a low-dimensional representation of the data.} This method overcomes the need to sweep over latent space dimensions. We observe that factoring out symmetries improves the MSE of the reconstruction as well as the dimension estimates and robustness to IRMAE-WD hyperparameters. When considering the original data (i.e. no symmetries factored out) and the phase-aligned data (only continuous symmetry factored out) the MSE is higher and the range of dimension estimates is wider. We also note that the dimension estimate range in the fundamental chart agrees well with the dimension found in our previous work for the same system \cite{perezde2023data}. 
We then train a NODE with latent space of dimension $d_h=10$ for both phase-aligned and fundamental domain data. This dimension is the upper bound of the dimension estimate obtained for the fundamental chart data. The resulting models using symmetry charting to map to the fundamental domain accurately reconstructed the DNS at both short and long times. In contrast, the phase-aligned model quickly landed on an unphysical stable RPO leading to poor reconstruction of the dynamics.

The approach described here is essentially a version of the ``CANDyMan" (Charts and Atlases for Nonlinear Data-driven Dynamics on Manifolds) approach described in \cite{floryan2021charts,fox2023predicting}.   
As shown in those studies, learning charts and atlases to develop \textit{local} manifold representations and dynamical models can improve performance for systems with complex dynamics. There, however, the charts were found by clustering; here that step is bypassed by using a priori knowledge, about the system, namely its symmetries. The methodology presented in this work can be applied to other dynamical systems with rich symmetries  by identifying the corresponding indicators. 
Future directions for symmetry charting and CANDyMan include applications such as control in systems with symmetry (cf. \cite{zeng2021symmetry}) as well as development of hierarchical methods where the fundamental chart identified by symmetry can be further subdivided using a cluster algorithm or a chart autoencoder \cite{Schonsheck.2019}. 

\begin{acknowledgments}
This work was supported by AFOSR  FA9550-18-1-0174 and ONR N00014-18-1-2865 (Vannevar Bush Faculty Fellowship). We also want to thank the Graduate Engineering Research Scholars (GERS) program and funding through the Advanced Opportunity Fellowship (AOF) as well as the PPG Fellowship.

\end{acknowledgments}


\begin{thebibliography}{}

\end{thebibliography}


\begin{thebibliography}{10}

\bibitem{srinivasan2019predictions}
Prem~A Srinivasan, L~Guastoni, Hossein Azizpour, Philipp Schlatter, and Ricardo Vinuesa.
\newblock Predictions of turbulent shear flows using deep neural networks.
\newblock {\em Physical Review Fluids}, 4(5):054603, 2019.

\bibitem{Page2020}
Jacob Page, Michael~P. Brenner, and Rich~R. Kerswell.
\newblock Revealing the state space of turbulence using machine learning.
\newblock {\em Phys. Rev. Fluids}, 6:034402, 2021.

\bibitem{doan2021auto}
Nguyen Anh~Khoa Doan, Wolfgang Polifke, and Luca Magri.
\newblock Auto-encoded reservoir computing for turbulence learning.
\newblock In {\em International Conference on Computational Science}, pages 344--351. Springer, 2021.

\bibitem{nakamura2021convolutional}
Taichi Nakamura, Kai Fukami, Kazuto Hasegawa, Yusuke Nabae, and Koji Fukagata.
\newblock Convolutional neural network and long short-term memory based reduced order surrogate for minimal turbulent channel flow.
\newblock {\em Physics of Fluids}, 33(2):025116, 2021.

\bibitem{foias1988modelling}
Ciprian Foias, O~Manley, and Roger Temam.
\newblock Modelling of the interaction of small and large eddies in two dimensional turbulent flows.
\newblock {\em ESAIM: Mathematical Modelling and Numerical Analysis}, 22(1):93--118, 1988.

\bibitem{temam1989inertial}
R~Temam.
\newblock Do inertial manifolds apply to turbulence?
\newblock {\em Physica D: Nonlinear Phenomena}, 37(1-3):146--152, 1989.

\bibitem{zelik2022attractors}
Sergey Zelik.
\newblock Attractors. {T}hen and now.
\newblock {\em arXiv preprint arXiv:2208.12101}, 2022.

\bibitem{linot2020deep}
Alec~J Linot and Michael~D Graham.
\newblock Deep learning to discover and predict dynamics on an inertial manifold.
\newblock {\em Physical Review E}, 101(6):062209, 2020.

\bibitem{linot2022data}
Alec~J Linot and Michael~D Graham.
\newblock Data-driven reduced-order modeling of spatiotemporal chaos with neural ordinary differential equations.
\newblock {\em Chaos: An Interdisciplinary Journal of Nonlinear Science}, 32(7):073110, 2022.

\bibitem{perezde2023data}
Carlos~E P{\'e}rez De~Jes{\'u}s and Michael~D Graham.
\newblock Data-driven low-dimensional dynamic model of {K}olmogorov flow.
\newblock {\em Physical Review Fluids}, 8(4):044402, 2023.

\bibitem{Linot.2023.10.1017/jfm.2023.720}
Alec~J. Linot and Michael~D. Graham.
\newblock {Dynamics of a data-driven low-dimensional model of turbulent minimal Couette flow}.
\newblock {\em Journal of Fluid Mechanics}, 973:A42, 2023.

\bibitem{Zeng.2024.10.1088/2632-2153/ad4ba5}
Kevin Zeng, Carlos~E P\'erez De~Jes\'us, Andrew~J Fox, and Michael~D Graham.
\newblock {Autoencoders for discovering manifold dimension and coordinates in data from complex dynamical systems}.
\newblock {\em Machine Learning: Science and Technology}, 2024.

\bibitem{jing2020implicit}
Li~Jing, Jure Zbontar, et~al.
\newblock Implicit rank-minimizing autoencoder.
\newblock {\em Advances in Neural Information Processing Systems}, 33:14736--14746, 2020.

\bibitem{lee2013smooth}
John~M Lee.
\newblock Smooth manifolds.
\newblock In {\em Introduction to smooth manifolds}, pages 1--31. Springer, 2013.

\bibitem{floryan2021charts}
Daniel Floryan and Michael~D Graham.
\newblock Data-driven discovery of intrinsic dynamics.
\newblock {\em Nature Machine Intelligence}, 4(12):1113--1120, 2022.

\bibitem{fox2023predicting}
Andrew~J Fox, C~Ricardo Constante-Amores, and Michael~D Graham.
\newblock Predicting extreme events in a data-driven model of turbulent shear flow using an atlas of charts.
\newblock {\em Physical Review Fluids}, 8(9):094401, 2023.

\bibitem{Budanur2017}
Nazmi~Burak Budanur and P.~Cvitanovi{\'{c}}.
\newblock {Unstable Manifolds of Relative Periodic Orbits in the Symmetry-Reduced State Space of the Kuramoto–Sivashinsky System}.
\newblock {\em Journal of Statistical Physics}, 167(3-4):636--655, 2017.

\bibitem{kneer2021symmetry}
Simon Kneer, Taraneh Sayadi, Denis Sipp, Peter Schmid, and Georgios Rigas.
\newblock Symmetry-{A}ware {A}utoencoders: s-{P}{C}{A} and s-nl{P}{C}{A}.
\newblock {\em arXiv preprint arXiv:2111.02893}, 2021.

\bibitem{miranda1993proto}
Rick Miranda and Emily Stone.
\newblock The proto-{L}orenz system.
\newblock {\em Physics Letters A}, 178(1-2):105--113, 1993.

\bibitem{zeng2021symmetry}
Kevin Zeng and Michael~D Graham.
\newblock Symmetry reduction for deep reinforcement learning active control of chaotic spatiotemporal dynamics.
\newblock {\em Physical Review E}, 104(1):014210, 2021.

\bibitem{meshalkin1961investigation}
LD~Meshalkin and Ia~G Sinai.
\newblock Investigation of the stability of a stationary solution of a system of equations for the plane movement of an incompressible viscous liquid.
\newblock {\em Journal of Applied Mathematics and Mechanics}, 25(6):1700--1705, 1961.

\bibitem{chandler2013invariant}
Gary~J Chandler and Rich~R Kerswell.
\newblock Invariant recurrent solutions embedded in a turbulent two-dimensional {K}olmogorov flow.
\newblock {\em Journal of Fluid Mechanics}, 722:554--595, 2013.

\bibitem{platt1991investigation}
Nathan Platt, L~Sirovich, and N~Fitzmaurice.
\newblock An investigation of chaotic {K}olmogorov flows.
\newblock {\em Physics of Fluids A: Fluid Dynamics}, 3(4):681--696, 1991.

\bibitem{armbruster1992phase}
Dieter Armbruster, Randy Heiland, Eric~J Kostelich, and Basil Nicolaenko.
\newblock Phase-space analysis of bursting behavior in {K}olmogorov flow.
\newblock {\em Physica D: Nonlinear Phenomena}, 58(1-4):392--401, 1992.

\bibitem{thess1992instabilities}
Andr{\'e} Thess.
\newblock Instabilities in two-dimensional spatially periodic flows. {P}art {I}: {K}olmogorov flow.
\newblock {\em Physics of Fluids A: Fluid Dynamics}, 4(7):1385--1395, 1992.

\bibitem{inubushi2012covariant}
Masanobu Inubushi, Miki~U Kobayashi, Shin-ichi Takehiro, and Michio Yamada.
\newblock Covariant {L}yapunov analysis of chaotic {K}olmogorov flows.
\newblock {\em Physical Review E}, 85(1):016331, 2012.

\bibitem{Farazmand2016}
M.~Farazmand.
\newblock {An adjoint-based approach for finding invariant solutions of Navier-Stokes equations}.
\newblock {\em Journal of Fluid Mechanics}, 795:278--312, 2016.

\bibitem{Suri.2020.10.1103/physrevlett.125.064501}
Balachandra Suri, Logan Kageorge, Roman~O. Grigoriev, and Michael~F. Schatz.
\newblock {Capturing Turbulent Dynamics and Statistics in Experiments with Unstable Periodic Orbits}.
\newblock {\em Physical Review Letters}, 125(6):064501, 2020.

\bibitem{crowley2022turbulence}
Christopher~J Crowley, Joshua~L Pughe-Sanford, Wesley Toler, Michael~C Krygier, Roman~O Grigoriev, and Michael~F Schatz.
\newblock Turbulence tracks recurrent solutions.
\newblock {\em Proceedings of the National Academy of Sciences}, 119(34):e2120665119, 2022.

\bibitem{armbruster1996symmetries}
D~Armbruster, B~Nicolaenko, N~Smaoui, and Pascal Chossat.
\newblock Symmetries and dynamics for 2-{D} {N}avier-{S}tokes flow.
\newblock {\em Physica D: Nonlinear Phenomena}, 95(1):81--93, 1996.

\bibitem{chen2018neural}
Ricky~TQ Chen, Yulia Rubanova, Jesse Bettencourt, and David~K Duvenaud.
\newblock Neural ordinary differential equations.
\newblock {\em Advances in neural information processing systems}, 31, 2018.

\bibitem{iudovich1965example}
VI~Iudovich.
\newblock Example of the generation of a secondary stationary or periodic flow when there is loss of stability of the laminar flow of a viscous incompressible fluid.
\newblock {\em Journal of Applied Mathematics and Mechanics}, 29(3):527--544, 1965.

\bibitem{green1974two}
JSA Green.
\newblock Two-dimensional turbulence near the viscous limit.
\newblock {\em Journal of Fluid Mechanics}, 62(2):273--287, 1974.

\bibitem{bartello1996self}
Peter Bartello and Tom Warn.
\newblock Self-similarity of decaying two-dimensional turbulence.
\newblock {\em Journal of Fluid Mechanics}, 326:357--372, 1996.

\bibitem{budanur2015periodic}
Nazmi~Burak Budanur, Daniel Borrero-Echeverry, and Predrag Cvitanovi{\'c}.
\newblock Periodic orbit analysis of a system with continuous symmetry---{A} tutorial.
\newblock {\em Chaos: An Interdisciplinary Journal of Nonlinear Science}, 25(7):073112, 2015.

\bibitem{budanur2015reduction}
Nazmi~Burak Budanur, Predrag Cvitanovi{\'c}, Ruslan~L Davidchack, and Evangelos Siminos.
\newblock Reduction of {S}{O} (2) symmetry for spatially extended dynamical systems.
\newblock {\em Physical review letters}, 114(8):084102, 2015.

\bibitem{linot2023stabilized}
Alec~J Linot, Joshua~W Burby, Qi~Tang, Prasanna Balaprakash, Michael~D Graham, and Romit Maulik.
\newblock Stabilized neural ordinary differential equations for long-time forecasting of dynamical systems.
\newblock {\em Journal of Computational Physics}, 474:111838, 2023.

\bibitem{Young.2023.10.1007/s00397-023-01412-0}
Charles~D. Young, Patrick~T. Corona, Anukta Datta, Matthew~E. Helgeson, and Michael~D. Graham.
\newblock {Scattering-Informed Microstructure Prediction during Lagrangian Evolution (SIMPLE)—a data-driven framework for modeling complex fluids in flow}.
\newblock {\em Rheologica Acta}, 62(10):587--604, 2023.

\bibitem{Kidger2021}
Patrick Kidger.
\newblock {\em On {Neural} {Differential} {Equations}}.
\newblock {PhD} {Thesis}, University of Oxford, 2021.

\bibitem{Finlay2020}
Chris Finlay, Joern-Henrik Jacobsen, Levon Nurbekyan, and Adam Oberman.
\newblock How to {Train} {Your} {Neural} {ODE}: the {World} of {Jacobian} and {Kinetic} {Regularization}.
\newblock In {\em Proceedings of the 37th {International} {Conference} on {Machine} {Learning}}, pages 3154--3164. PMLR, November 2020.
\newblock ISSN: 2640-3498.

\bibitem{Kelly2020}
Jacob Kelly, Jesse Bettencourt, Matthew~J Johnson, and David~K Duvenaud.
\newblock Learning {Differential} {Equations} that are {Easy} to {Solve}.
\newblock In {\em Advances in {Neural} {Information} {Processing} {Systems}}, volume~33, pages 4370--4380. Curran Associates, Inc., 2020.

\bibitem{Mezić.2021.10.1090/noti2306}
Igor Mezić.
\newblock {Koopman Operator, Geometry, and Learning of Dynamical Systems}.
\newblock {\em Notices of the American Mathematical Society}, 68(07):1, 2021.

\bibitem{Brunton.2022.10.1137/21m1401243}
Steven~L Brunton, Marko Budišić, Eurika Kaiser, and J~Nathan Kutz.
\newblock {Modern Koopman Theory for Dynamical Systems}.
\newblock {\em SIAM Review}, 64(2):229--340, 2022.

\bibitem{Buzhardt.2024}
Jake Buzhardt, C~Ricardo Constante-Amores, and Michael~D Graham.
\newblock {On the relationship between Koopman operator approximations and neural ordinary differential equations for data-driven time-evolution predictions}.
\newblock {\em arXiv}, 2024.

\bibitem{boser1992training}
Bernhard~E Boser, Isabelle~M Guyon, and Vladimir~N Vapnik.
\newblock A training algorithm for optimal margin classifiers.
\newblock In {\em Proceedings of the fifth annual workshop on Computational learning theory}, pages 144--152, 1992.

\bibitem{loshchilov2017decoupled}
Ilya Loshchilov and Frank Hutter.
\newblock Decoupled weight decay regularization.
\newblock {\em arXiv preprint arXiv:1711.05101}, 2017.

\bibitem{Schonsheck.2019}
Stefan Schonscheck, Jie Chen, and Rongjie Lai.
\newblock {Chart Auto-Encoders for Manifold Structured Data}.
\newblock {\em arXiv}, 2019.

\bibitem{Gibson2008}
J.~F. Gibson, J.~Halcrow, and P.~Cvitanovi{\'{c}}.
\newblock {Visualizing the geometry of state space in plane {C}ouette flow}.
\newblock {\em Journal of Fluid Mechanics}, 611(1987):107--130, 2008.

\bibitem{Aghor2024}
Pratik~P. Aghor and John~F. Gibson.
\newblock Symmetry groups and invariant solutions of plane {P}oiseuille flow.
\newblock {\em arXiv preprint arXiv:2409.11517}, 2024.

\end{thebibliography}

\appendix  
\section{Symmetry operations in Fourier space}\label{sec:AppendixA}
In this appendix we show the derivations to obtain the actions of the symmetry operations on the Fourier coefficients as shown in Equations \ref{eq:SR_fourier_c} through \ref{eq:shift_fourier_c}. We start with the shift (in $y$)-reflect (in $x$) symmetry
\begin{gather}
\mathscr{S}:[u, v, \omega](x, y) \rightarrow[-u, v,-\omega]\left(-x, y+\frac{\pi}{n}\right).
\end{gather}
Throughout this appendix we will focus on the vorticity representation $\omega(x,y)$\MDGrevise{, which we can write in terms of its Fourier coefficients $\hat{\omega}(k_x,k_y)$ as} 

\begin{gather}
\omega(x,y) = \sum_{k_x}\sum_{k_y} \hat{\omega}(k_x,k_y) e^{i(k_x x + k_y y)}.
\end{gather}
The symmetry operations act term-by-term on the Fourier series. Applying the shift (in $y$)-reflect (in $x$) symmetry to each term yields
\begin{gather}
\mathscr{S}:\hat{\omega}(k_x, k_y)e^{i(k_x x + k_y y)} \rightarrow -\hat{\omega}(k_x, k_y)e^{-ik_x x + ik_y y + ik_y\frac{\pi}{n}}.
\end{gather}
We can simplify the equation to get 
\begin{gather}
\mathscr{S}:\hat{\omega}(k_x, k_y) \rightarrow -\hat{\omega}(-k_x, k_y)e^{ik_y\frac{\pi}{n}}.
\end{gather}
For the rotation through $\pi$,
\begin{gather}
\mathscr{R}:\omega(x, y) \rightarrow \omega(-x,-y), 
\end{gather}
we can also write in Fourier representation
\begin{gather}
\mathscr{R}:\hat{\omega}(k_x, k_y)e^{i(k_x x + k_y y)} \rightarrow \hat{\omega}(k_x, k_y)e^{-i(k_x x + ik_y y)}.
\end{gather}
Simplifying this equation we get
\begin{gather}
\mathscr{R}:\hat{\omega}(k_x, k_y) \rightarrow \hat{\omega}(-k_x, -k_y).
\end{gather}
Since $\hat{\omega}(k_x, k_y)$ and  $\hat{\omega}(-k_x, -k_y)$ are complex conjugates, we can write this as 
\begin{gather}
\mathscr{R}:\hat{\omega}(k_x, k_y) \rightarrow \bar{\hat{\omega}}(k_x, k_y).
\end{gather}
For the continuous translation in $x$,
\begin{gather}
\mathscr{T}_{l}: \omega (x, y) \rightarrow \omega(x+l, y),
\end{gather}
we can write in Fourier representation 
\begin{gather}
\mathscr{T}_{l}:\hat{\omega}(k_x, k_y)e^{i(k_x x + k_y y)} \rightarrow \hat{\omega}(k_x, k_y)e^{ik_x x + ik_y y + ik_x l)}.
\end{gather}
Simplifying this equation we get
\begin{gather} 
	\mathscr{T}_l:\hat{\omega}(k_x, k_y) \rightarrow \hat{\omega}(k_x, k_y)e^{i k_x l}.
\end{gather}


\section{Symmetry operations and indicators for plane Couette and Poiseuille  flows}\label{sec:channelAppendix}

Here we show how to select a set of  indicators for plane Couette flow (PCF) and plane Poiseuille flow (PPF) with periodic boundary conditions in $x$ and $z$ and no-slip boundary conditions at $y=-1$ and $y=1$. For PCF, the Navier-Stokes equations and boundary conditions are invariant to two discrete symmetry operations ($\sigma_{C1}$, $\sigma_{C2}$) and a continuous translation $\mathscr{T}_{C}(l_x,l_z)$ \citep{Gibson2008}:
\begin{gather}
\sigma_{C1}:[u, v, w](x, y,z) \rightarrow[u, v,-w](x,y,-z), \label{eq:sig1PCF}\\
\sigma_{C2}:[u, v, w](x,y,z) \rightarrow[-u,-v, w](-x,-y,z), \label{eq:sig2PCF}\\
\mathscr{T}_{C}(l_x,l_z):[u, v, w](x, y,z) \rightarrow[u, v, w](x+l_x, y,z+l_z).\label{eq:transPCF}
\end{gather}
\MDGrevise{The operation $\sigma_{C1}$ corresponds to reflecting the velocity field across the $x-z$ plane, and $\sigma_{C2}$ to rotating it around the $x$ axis.}
Similarly, for PPF, the Navier-Stokes equations and boundary conditions are invariant to two discrete symmetries ($\sigma_{P1}$, $\sigma_{P2}$) and a continuous translation $\mathscr{T}_{P}(l_x,l_z)$ \citep{Aghor2024}:
\begin{gather}
\sigma_{P1}:[u, v, w](x, y,z) \rightarrow[u, v,-w](x,y,-z), \label{eq:sig1PPF}\\
\sigma_{P2}:[u, v, w](x,y,z) \rightarrow[u,-v, w](x,-y,z), \label{eq:sig2PPF}\\
\mathscr{T}_{P}(l_x,l_z):[u, v, w](x, y,z) \rightarrow[u, v, w](x+l_x, y,z+l_z).\label{eq:transPPF}
\end{gather}
Note that $\sigma_{C1}=\sigma_{P1}$ and $\mathscr{T}_C(l_x,l_z)=\mathscr{T}_P(l_x,l_z)$, while $\sigma_{C2}$ and $\sigma_{P2}$ differ in whether $u$ changes sign (Poiseuille) or not (Couette) under rotation. 

The translation symmetries in $x$ and $z$ can be treated just as we did for Kolmogorov flow, using the first Fourier-mode method of slices in $x$ and $z$. Since there are two discrete symmetry operations, each data point lies in one of four discrete subspaces, and our goal is to find criteria that uniquely identify which one. 

For PCF and PPF, it is useful to consider the action of these symmetry operations on the Fourier-Chebyshev representation of the flow, given by
\begin{gather}
\mathbf{u}(x,y,z) = \sum_{k_x}\sum_{k_y}\sum_{k_z} \hat{\mathbf{u}}(k_x,k_y,k_z) T_{k_y}(y)e^{i(k_x x + k_z z)},
\end{gather}
where $\mathbf{u}=[u,v,w]$, $\hat{\mathbf{u}}=[\hat{u},\hat{v},\hat{w}]$, and $T_{k_y}(y)$ are Chebyshev polynomials. Under this transformation, the symmetry operations are 
\begin{gather}
\sigma_{C1}:[\hat{u}, \hat{v}, \hat{w}](k_x, k_y,k_z) \rightarrow[\hat{u}, \hat{v},-\hat{w}](k_x,k_y,-k_z), \label{eq:sig1fft}\\
\sigma_{C2}:[\hat{u}, \hat{v}, \hat{w}](k_x, k_y,k_z) \rightarrow(-1)^{k_y}[-\hat{u},-\hat{v}, \hat{w}](-k_x,k_y,k_z), \label{eq:sig2fft}\\
\sigma_{P2}:[\hat{u}, \hat{v}, \hat{w}](k_x, k_y,k_z) \rightarrow(-1)^{k_y}[\hat{u},-\hat{v}, \hat{w}](k_x,k_y,k_z), \label{eq:sig3fft}\\
\mathscr{T}_{C}(l_x,l_z):[\hat{u}, \hat{v}, \hat{w}](k_x, k_y,k_z) \rightarrow[\hat{u}, \hat{v}, \hat{w}](k_x, k_y,k_z)e^{i(k_xl_x+k_zl_z)}.\label{eq:transfft}
\end{gather}

Now, we can consider specific values of $(k_x,k_y,k_z)$ to select our criteria.  Basing the criteria on $k_x=0$ and $k_z=0$ coefficients decouples them from phase-alignment with the method-of-slices. Additionally, we can use the fact that $\sigma_{C2}$ and $\sigma_{P2}$ modify $\hat{v}(0,k_y,0)$ and $\hat{w}(0,k_y,0)$ in the same way to identify the same criteria for both PCF and PPF. We do this by considering the effect of the discrete symmetry operations on $\hat{v}(0,2,0)$:
\begin{gather}
\sigma_{C1},\sigma_{P1}:\hat{v}(0,2,0) \rightarrow \hat{v}(0,2,0),\\
\sigma_{C2}:\hat{v}(0,2,0) \rightarrow -\hat{v}(0,2,0),\\
\sigma_{P2}:\hat{v}(0,2,0) \rightarrow -\hat{v}(0,2,0),
\end{gather}
and on $\hat{w}(0,2,0)$:
\begin{gather}
\sigma_{C1},\sigma_{P1}:\hat{w}(0,2,0) \rightarrow -\hat{w}(0,2,0),\\
\sigma_{C2}:\hat{w}(0,2,0) \rightarrow \hat{w}(0,2,0),\\
\sigma_{P2}:\hat{w}(0,2,0) \rightarrow \hat{w}(0,2,0).
\end{gather}

\MDGrevise{Observe that $\sigma_{C1}(=\sigma_{P1})$ changes the sign of $\hat{w}(0,2,0) $ but not $\hat{v}(0,2,0)$, so we use the former as a criterion for this operation. For  $\sigma_{C2}$ and $\sigma_{P2}$, the opposite occurs, so we use the sign of $\hat{v}(0,2,0)$ as the criterion for those operators.}
In summary, we can define criteria, and corresponding integer indicators $\mathcal{I}$, based on the quadrant in ($\hat{v}(0,2,0)$,$\hat{w}(0,2,0)$) in which the data lies, as shown in Table \ref{table_AppB}.

\begin{table}[t]
\caption{Indicators $\mathcal{I}$, corresponding operations required to map to $\mathcal{I}=0$, the fundamental domain, and signs of the relevant Fourier-Chebyshev coefficients, for PCF and PPF.}
$$
\begin{array}{cccc}\hline \hline \mathcal{I} & \text {\; \; \; \; Discrete operation } & \sgn(\hat{v}(0,2,0)) & \sgn(\hat{w}(0,2,0))\\ \hline  0 & \text{Identity} & + & +  \\ 
1 & \sigma_{C1},\sigma_{P1} & + & - \\  
2 & \sigma_{C2},\sigma_{P2} & - & + \\  
3 & \sigma_{C1}\sigma_{C2},\sigma_{P1}\sigma_{P2} & - & - \\ \hline \hline\end{array}
$$
\label{table_AppB}
\end{table}

\end{document}